\documentclass[utf8]{frontiersSCNS} 

\usepackage{url,hyperref,lineno,microtype,subcaption}
\usepackage[onehalfspacing]{setspace}

\usepackage{amsmath,amsfonts}
\usepackage{algorithm}
\usepackage{algorithmic}
\usepackage{graphicx}
\usepackage{textcomp}
\usepackage{xcolor}
\usepackage{booktabs}
\usepackage{multirow}
\usepackage[super]{nth}
\usepackage{tikz}

\newcommand*\circled[1]{\tikz[baseline=(char.base)]{
		\node[shape=circle,draw,inner sep=0.2pt] (char){#1};}}
\newcommand*\circledB[1]{\tikz[baseline=(char.base)]{
            \node[shape=circle,fill,inner sep=0.2pt] (char) {\textcolor{white}{#1}};}}
\usepackage{soul}
\usepackage{setspace}
\newcommand{\minus}{\scalebox{0.4}[1.0]{$-$}}
\usepackage{fancyhdr}
\fancypagestyle{firstpage}
{
  \fancyhf{}
  \chead{\small Accepted for publication at Frontiers in Neuroscience - Section Neuromorphic Engineering.}
  \cfoot{\thepage}
}

\def\keyFont{\fontsize{8}{11}\helveticabold }
\def\firstAuthorLast{Rachmad Vidya Wicaksana Putra {et~al.}} 

\def\Authors{Rachmad Vidya Wicaksana Putra\,$^{1,*}$, Muhammad Abdullah Hanif\,$^{2}$ and Muhammad Shafique\,$^{2}$}

\def\Address{$^{1}$ Embedded Computing Systems (ECS), Institute of Computer Engineering, Technische Universit\"at Wien (TU Wien), Vienna, Austria \\
$^{2}$ eBrain Lab, Division of Engineering, New York University Abu Dhabi (NYUAD), Abu Dhabi, United Arab Emirates  
}

\def\corrAuthor{Rachmad Vidya Wicaksana Putra}
\def\corrEmail{rachmad.putra@tuwien.ac.at}

\begin{document}
\onecolumn
\firstpage{1}

\title[EnforceSNN: SNNs with Approximate DRAMs]{EnforceSNN: Enabling Resilient and Energy-Efficient Spiking Neural Network Inference considering Approximate DRAMs for Embedded Systems} 

\author[\firstAuthorLast]{\Authors} %This field will be automatically populated
\address{} %This field will be automatically populated
\correspondance{} %This field will be automatically populated

\extraAuth{}% If there are more than 1 corresponding author, comment this line and uncomment the next one.
%\extraAuth{corresponding Author2 \\ Laboratory X2, Institute X2, Department X2, Organization X2, Street X2, City X2 , State XX2 (only USA, Canada and Australia), Zip Code2, X2 Country X2, email2@uni2.edu}

\maketitle
\thispagestyle{firstpage}

%%%%%%%%%%%%%%%%%%%%%%%%%%%%%%%%%%%%%%%%%%%%%%%%%%%%%%%
%%%%%%%%%%%%%%%%%%%%%%%%%%%%%%%%%%%%%%%%%%%%%%%%%%%%%%%

\vspace{0.5cm}
\begin{abstract}
Spiking Neural Networks (SNNs) have shown capabilities of achieving high accuracy under unsupervised settings and low operational power/energy due to their bio-plausible computations.
Previous studies identified that DRAM-based off-chip memory accesses dominate the energy consumption of SNN processing.
However, state-of-the-art works do not optimize the DRAM energy-per-access, thereby hindering the SNN-based systems from achieving further energy efficiency gains. 
To substantially reduce the DRAM energy-per-access, an effective solution is to decrease the DRAM supply voltage, but it may lead to errors in DRAM cells (i.e., so-called \textit{approximate DRAM}).
Towards this, we propose \textit{EnforceSNN}, a novel design framework that provides a solution for resilient and energy-efficient SNN inference using reduced-voltage DRAM for embedded systems. 
The key mechanisms of our EnforceSNN are: 
(1) employing quantized weights to reduce the DRAM access energy; 
(2) devising an efficient DRAM mapping policy to minimize the DRAM energy-per-access;
(3) analyzing the SNN error tolerance to understand its accuracy profile considering different bit error rate (BER) values;  
(4) leveraging the information for developing an efficient fault-aware training (FAT) that considers different BER values and bit error locations in DRAM to improve the SNN error tolerance; and 
(5) developing an algorithm to select the SNN model that offers good trade-offs among accuracy, memory, and energy consumption.
The experimental results show that our EnforceSNN maintains the accuracy (i.e., no accuracy loss for $BER \leq 10^{-3}$) as compared to the baseline SNN with accurate DRAM, while achieving up to 84.9\% of DRAM energy saving and up to 4.1x speed-up of DRAM data throughput across different network sizes.

\tiny
\keyFont{\section{Keywords:} spiking neural networks, high performance, energy efficiency, approximate computing, approximate DRAM, DRAM errors, error tolerance, resilience.} 
\end{abstract}

%%%%%%%%%%%%%%%%%%%%%%%%%%%%%%%%%%%%%%%%%%%%%%%%%%%%%%%
%%%%%%%%%%%%%%%%%%%%%%%%%%%%%%%%%%%%%%%%%%%%%%%%%%%%%%%
\section{Introduction}
\label{Sec_Intro}

Spiking neural networks (SNNs) have demonstrated the potential of obtaining high accuracy under unsupervised settings and low operational energy due to their bio-plausible spike-based computations~\citep{Ref_Putra_FSpiNN_TCAD20}.
A larger SNN model is usually favorable as it offers higher accuracy than the smaller ones, as shown by our experimental results in Fig.~\ref{Fig_ObserveSNNnDRAM}(a). 
Here, the 1MB-sized network achieves only 75\%, while the 200MB-sized network achieves 92\% accuracy for the MNIST dataset. 
This MNIST dataset is a set of training and test images for handwritten digit 0-9~\citep{Ref_Lecun_MNIST_IEEE98}.
On the other hand, most of the SNN hardware platforms have relatively small on-chip memory, e.g., less than 100MB~\citep{Ref_Roy_PEASE_ISLPED17,Ref_Sen_ApproxSNN_DATE17,Ref_Frenkel_ODIN_TBCAS19,Ref_Frenkel_MorphIC_TBCAS19}. 
Therefore, running an SNN model with a larger size than the on-chip memory of SNN hardware platforms, will require intensive accesses to the off-chip memory.
Previous studies show that single access to the off-chip memory (i.e., DRAM) incurs significantly higher energy consumption than single access to the on-chip memory (i.e., SRAM)~\cite{Ref_Sze_DNNsurvey_IEEE17,Ref_Putra_ROMANet_TVLSI21}.
Moreover, previous work also identified that memory accesses dominate the energy consumption of SNN processing, incurring 50\%-75\% of the total system energy across different SNN hardware platforms, as shown in Fig.~\ref{Fig_ObserveSNNnDRAM}(b). 
The reason is that, DRAM access energy is significantly higher than other SNN operations (e.g., neuron operations)~\citep{Ref_Krithivasan_SpikeBundle_ISLPED19}.
This problem is even more critical for AI applications with stringent constraints (e.g., low-cost embedded devices with a small on-chip memory size)~\citep{Ref_Shafique_EdgeAI_ICCAD21}, since it leads to even more intensive DRAM accesses.
Consequently, this problem hinders SNN-based embedded systems from obtaining further efficiency gains.

\textbf{Targeted Research Problem:} 
\textit{How can we substantially decrease the DRAM access energy for the SNN inference, while maintaining the accuracy}. 
The solution to this problem will enable efficient SNN inference for energy-constrained embedded devices and their applications for Edge-AI and Smart CPS. 
\textit{Edge-AI} is the system that runs Artificial Intelligence (AI) algorithms on resource- and energy-constrained computing devices at the edge of the network, i.e., close to the source of data~\citep{Ref_Chen_DLwithEdge_JPROC19, Ref_Shi_Edge_JIOT16, Ref_Satyanarayanan_Edge_MC17, Ref_Yu_Edge_Access18, Ref_Liu_Edge_JPROC19, Ref_Cao_Edge_Access20}.
Meanwhile, \textit{Smart CPS (Cyber-Physical System)} is the system that includes the interacting networks of computational components (e.g., computation and storage devices), physical components (e.g., sensors and actuators), and human users~\citep{Ref_Griffor_CPS_NIST17, Ref_Chattopadhyay_CPS_DATE17, Ref_Shafique_SmartCPS_DSD18, Ref_Kriebel_CPSIoT_ISVLSI18}.

%%%%%%%%%%%%%%%%%%%%%%%
\subsection{State-of-the-Art and Their Limitations}
\label{Sec_Intro_SoA}

To decrease the energy consumption of SNN inference, state-of-the-art works have developed different techniques, which can be loosely classified as the following.
\begin{itemize}
    \item \textbf{Reduction of the SNN operations} through approximate neuron operations~\citep{Ref_Sen_ApproxSNN_DATE17}, weight pruning~\citep{Ref_Rathi_PruneQuantizeSNN_TCAD18}, and neuron removal~\citep{Ref_Putra_FSpiNN_TCAD20}. 
    These techniques decrease the number of DRAM accesses for the corresponding model parameters.
    \item \textbf{Quantization} by reducing the range of representable values for SNN parameters (e.g., weights)~\citep{Ref_Rathi_PruneQuantizeSNN_TCAD18,Ref_Putra_FSpiNN_TCAD20,Ref_Putra_QSpiNN_IJCNN21}. 
    These techniques reduce the amount of SNN parameters (e.g., weights) to be stored in and fetched from DRAM.
\end{itemize}

\begin{figure}[t]
\centering
\includegraphics[width=\linewidth]{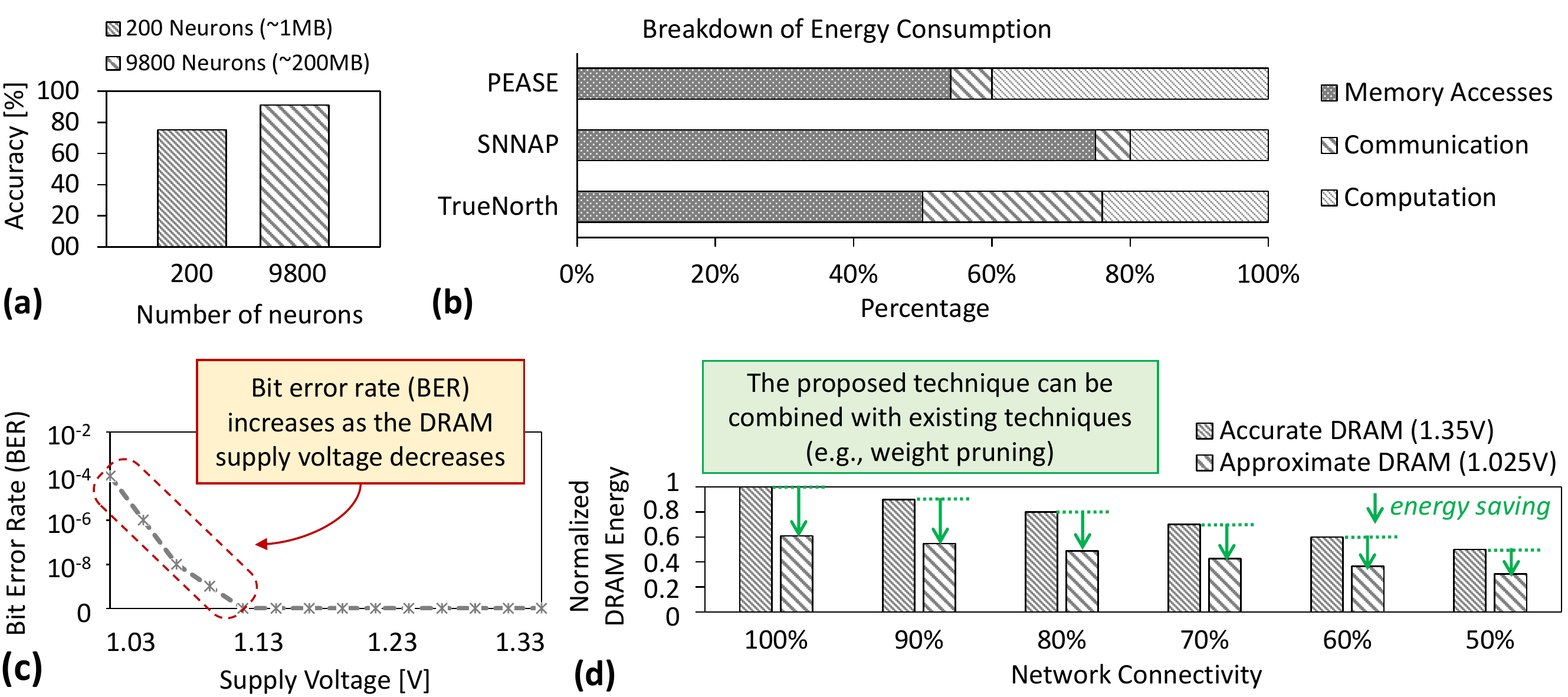}
\caption{(a) Accuracy profiles of small-sized and large-sized SNN models on the MNIST dataset which are obtained from our experiments. (b) Breakdown of the energy consumption of SNN processing on different SNN hardware platforms, i.e., PEASE~\citep{Ref_Roy_PEASE_ISLPED17}, SNNAP~\citep{Ref_Sen_ApproxSNN_DATE17}, and TrueNorth~\citep{Ref_Akopyan_TrueNorth_TCAD15}; adapted from studies in~\citep{Ref_Krithivasan_SpikeBundle_ISLPED19}. (c) Bit error rate (BER) of approximate DRAM and its respective supply voltage $V_{supply}$; adapted from studies in~\citep{Ref_Chang_Voltron_POMACS17}. (d) The estimated DRAM energy savings achieved by our technique when combined with the weight pruning across different rates of network connectivity (i.e., synaptic connections) for a network with 4900 excitatory neurons.
The results are obtained from experiments using the
LPDDR3-1600 4Gb DRAM configuration and the DRAMPower simulator~\citep{Ref_Chandrasekar_DRAMPower}.}
\label{Fig_ObserveSNNnDRAM}
\end{figure}

\textbf{Limitations:} 
These state-of-the-art works mainly aim at reducing the number of DRAM accesses, but do not optimize the DRAM energy-per-access and do not employ  approximations in DRAM that provide an additional knob for obtaining high energy efficiency.
Therefore, optimization gains offered by these works are sub-optimal, hindering the SNN inference systems from achieving the full potential of DRAM energy savings.
Therefore, the effective optimization should jointly minimize the DRAM energy-per-access and the number of DRAM accesses, by leveraging the approximation in DRAM to expose the full energy-saving potential, while overcoming the negative impact of the approximation-induced errors (i.e., bit-flips in DRAM cells). 
Fig.~\ref{Fig_ObserveSNNnDRAM}(c) shows the approximation-induced error rates in DRAM.

To address these limitations, \textit{we employ approximate DRAM (i.e., DRAM with reduced supply voltage) with efficient DRAM data mapping policy and fault-aware training to substantially reduce the DRAM access energy in SNN inference systems while preserving their accuracy}.
Moreover, our proposed technique can also be combined with state-of-the-art techniques to further improve the energy efficiency of SNN inference. 
For example, Fig.~\ref{Fig_ObserveSNNnDRAM}(d) shows the estimated DRAM energy savings achieved by our technique when combined with the weight pruning.
To highlight the potential of reduced-voltage approximate DRAM, we perform an experimental case study in the following section.

%%%%%%%%%%%%%%%%%%%%%%%
\subsection{Motivational Case Study and Key Challenges}
\label{Sec_Intro_CaseStudy}

\begin{figure}[t]
\centering
\includegraphics[width=\linewidth]{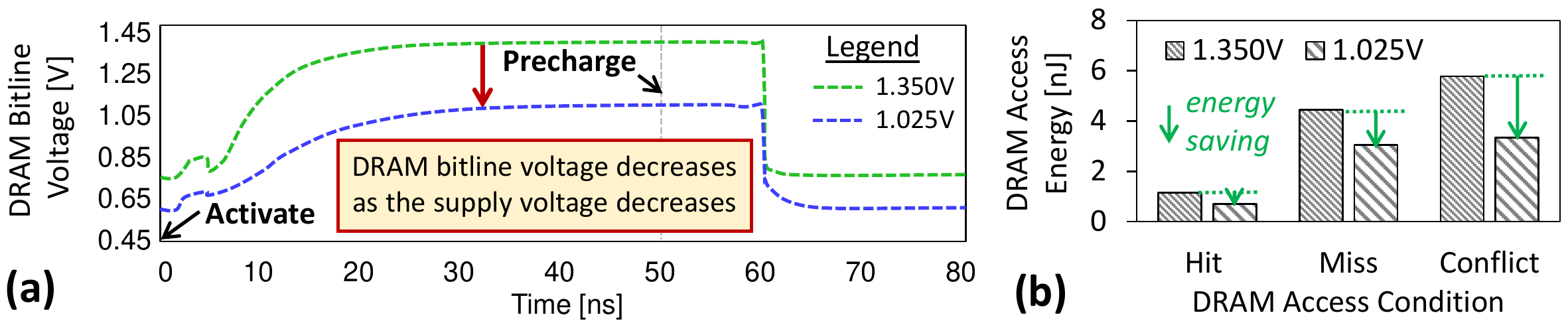}
\caption{(a) The dynamics of $V_{bitline}$ under different $V_{supply}$ values. (b) DRAM access energy for a row buffer hit, a row buffer miss, and a row buffer conflict, under different $V_{supply}$ values.}
\label{Fig_ObserveApproxDRAM}
\end{figure}

In the case study, we aim at studying (1) the dynamics of DRAM bitline voltage ($V_{bitline}$) for both the accurate and approximate DRAM settings, and (2) the DRAM access energy for different access conditions (including a row buffer hit, miss, or conflict). 
Note, $V_{bitline}$ is defined as the voltage measured in each DRAM bitline when a DRAM supply voltage ($V_{supply}$) is applied, as shown in Fig.~\ref{Fig_ObserveApproxDRAM}(a) and~\ref{Fig_DRAMnSpiceDRAM}(c). 
Further details on the dynamics of $V_{bitline}$ are provided in Section~\ref{Sec_Prelim_xDRAMvolt}.
For DRAM access conditions, a row buffer hit means that the requested data has been loaded in the DRAM row buffer, thus the data can be accessed without additional DRAM operations. 
Meanwhile, a row buffer miss or conflict needs to open the requested DRAM row before the data can be loaded into the row buffer and then accessed. 
Further information on the DRAM access conditions is discussed in Section~\ref{Sec_Prelim_xDRAMfund}.

For the experimental setup, we employ the DRAM circuit model from the work of~\citet{Ref_Chang_Voltron_POMACS17} and the SPICE simulator to study the dynamics of $V_{bitline}$.
The accurate DRAM operates at 1.35V of the supply voltage ($V_{supply}$), while the approximate one operates at 1.025V. 
Further details on the experimental setup are discussed in Section~\ref{Sec_Eval}. 
Furthermore, we consider the LPDDR3-1600 4Gb DRAM configuration as it is representative of the low-power DRAM types for embedded systems. 
We employ the DRAMPower simulator to estimate the DRAM access energy because it has been validated against real measurements~\citep{Ref_Chandrasekar_DRAMPower} and has been widely used in the computer architecture communities.
Fig.~\ref{Fig_ObserveApproxDRAM} presents the experimental results, from which we make the following key observations.
\begin{itemize}
    \item The $V_{bitline}$ decreases as the $V_{supply}$ decreases, hence forcing the DRAM cells to operate under lower reliability as the \textit{weak cells} may fail to hold the correct bits.
    \textit{Weak cells} are DRAM cells that fail when the DRAM parameters (e.g., voltage, timing) are reduced~\citep{Ref_Chang_Voltron_POMACS17,Ref_Kim_SolarDRAM_ICCD18}.
    \item The reduced-voltage DRAM decreases the DRAM energy-per-access across different access conditions, i.e., by up to 42\% of energy reduction for each access.
    \item The row buffer hit has lower energy consumption than the row buffer miss or conflict. 
    Moreover, row buffer hit also incurs less latency than the row buffer miss or conflict~\citep{Ref_Putra_DRMap_DAC20,Ref_Putra_ROMANet_TVLSI21}.
    Therefore, the row buffer hit should be exploited to optimize the DRAM latency and energy.
\end{itemize}

Although employing the approximate DRAM can substantially decrease the DRAM energy-per-access, it also decreases the DRAM reliability since the bit errors increase when the $V_{supply}$ decreases, as shown in Fig.~\ref{Fig_ObserveSNNnDRAM}(c).
These errors may degrade the accuracy of SNN inference since they can change the weight values in DRAM, which then deteriorates the neuron behavior. 

\textbf{Associated Research Challenge:} 
\textit{How to achieve low DRAM access energy for SNN inference using approximate DRAM, while minimizing their negative impact on the accuracy}.

%%%%%%%%%%%%%%%%%%%%%%%
\subsection{Our Novel Contributions}
\label{Sec_Intro_NovelContrib}

To overcome the above research challenges, we propose \textbf{the EnforceSNN framework}, which enables resilient and energy-efficient SNNs considering approximate DRAMs (i.e., reduced-voltage DRAMs) for embedded systems. 
Based on the best of our knowledge, it is the first effort that employs approximate DRAM for improving the energy efficiency of SNN inference, while enhancing their error tolerance against bit errors in DRAM. 
Our EnforceSNN framework employs the following key steps. 
\begin{enumerate}
    \item \textbf{Employing weight quantization} to reduce the memory footprint for SNN weights and the number of DRAM accesses for SNN inference, thereby optimizing the DRAM access energy.
    \item \textbf{Devising an efficient DRAM data mapping} to maximize row buffer hits for optimizing the DRAM energy-per-access while considering BER in DRAM.
    \item \textbf{Analyzing the SNN error tolerance} to understand the SNN accuracy profile under different DRAM supply voltage and different BER values. 
    \item \textbf{Improving the SNN error tolerance} by developing and employing efficient fault-aware training (FAT) that considers SNN accuracy profile and bit error locations in DRAM.
    \item \textbf{Devising an algorithm to select the SNN model} that offers good trade-offs among accuracy, memory, and energy consumption from the given model candidates using the proposed reward function.
\end{enumerate}

\textbf{Key Results:} 
We evaluate the EnforceSNN framework for (1) classification accuracy using PyTorch-based simulations~\citep{Ref_Hazan_BindsNET_FNINF18} on a multi-GPU machine considering the MNIST and Fashion MNIST datasets\footnote{The research works for the unsupervised learning-based SNNs is still in the early stage and typically use small datasets like the MNIST and the Fashion MNIST. We also adopt the same test conditions as used widely by the SNN research community.}, and (2) DRAM access energy using DRAMPower~\citep{Ref_Chandrasekar_DRAMPower}. 
We perform an epoch of unsupervised learning (60K experiments) for each retraining process considering each combination of the SNN model, workload, and training BER; then perform inference (10K experiments) for each combination of the SNN model, workload, and testing BER. 
The experimental results indicate that, our EnforceSNN reduces the DRAM access energy by up to 84.9\% and improves the speed-up up to 4.1x, while maintaining the accuracy (no accuracy loss) across different network sizes for $BER \leq 10^{-3}$.

%%%%%%%%%%%%%%%%%%%%%%%%%%%%%%%%%%%%%%%%%%%%%%%%%%%%%%%
%%%%%%%%%%%%%%%%%%%%%%%%%%%%%%%%%%%%%%%%%%%%%%%%%%%%%%%
\section{Background}
\label{Sec_Prelim}

%%%%%%%%%%%%%%%%%%%%%%%
\subsection{Spiking Neural Networks (SNNs)}
\label{Sec_Prelim_SNNs}

\begin{figure}[t]
\centering
\includegraphics[width=\linewidth]{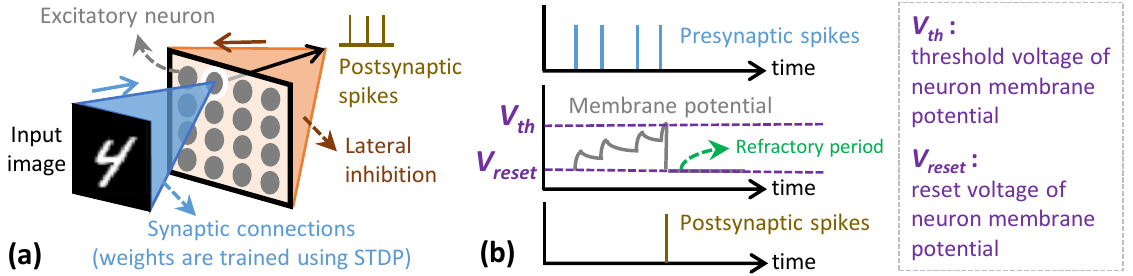}
\caption{(a) The SNN architecture considered in this work, which is adapted from~\citep{Ref_Putra_FSpiNN_TCAD20}. (b) The neuronal dynamics of the Leaky Integrate-and-Fire (LIF) neuron model.}
\label{Fig_SNN_n_LIF}
\end{figure}

SNNs are the neural network models that employ bio-plausible computations and use the sequences of spikes (i.e., \textit{spike trains}) for conveying information.
These spikes are encoded using a specific spike coding.
Several spike coding schemes have been proposed in the literature~\citep{Ref_Gautrais_SpikeCoding_Bio98, Ref_Thorpe_RankOrder_Springer98, Ref_Kayser_PhaseCoding_Neuron09, Ref_Park_BurstSNN_DAC19, Ref_Park_T2FSNN_DAC20}. 
Here, we use rate coding as it has been used widely and offers robustness for diverse learning rules~\citep{Ref_Diehl_STDPmnist_FNCOM15,Ref_Putra_ReSpawn_ICCAD21}.
For the learning rule, we use the spike-timing-dependent plasticity (STDP), as it has been widely used by previous works~\citep{Ref_Diehl_STDPmnist_FNCOM15,Ref_Saunders_STDPpatch_IJCNN18,Ref_Hazan_SOMSNN_IJCNN18,Ref_Putra_SpikeDyn_DAC21}.
In an SNN model, neurons and synapses are connected in a specific architecture~\citep{Ref_Pfeiffer_DLSNN_FNINS18,Ref_Tavanaei_DLSNN_Neunet18,Ref_Mozafari_SpykeTorch_FNINS19}. 
Here, we consider a fully-connected architecture as it supports unsupervised learning scenarios; see Fig.~\ref{Fig_SNN_n_LIF}(a). 
In this architecture, each input pixel is connected to all (excitatory) neurons, and the output of each neuron is connected to other neurons for providing inhibition.
For the neuron model, we use the Leaky Integrate-and-Fire (LIF) as it provides bio-plausible neuronal dynamics with low computational complexity~\citep{Ref_Izhikevich_CompareModels_TNN04,Ref_Putra_SoftSNN_arXiv22}; see Fig.~\ref{Fig_SNN_n_LIF}(b).

%%%%%%%%%%%%%%%%%%%%%%%
\subsection{Approximate DRAM}
\label{Sec_Prelim_xDRAM}

%%%%%%%%%%
\subsubsection{DRAM Fundamentals}
\label{Sec_Prelim_xDRAMfund}

\begin{figure}[t]
\centering
\includegraphics[width=0.8\linewidth]{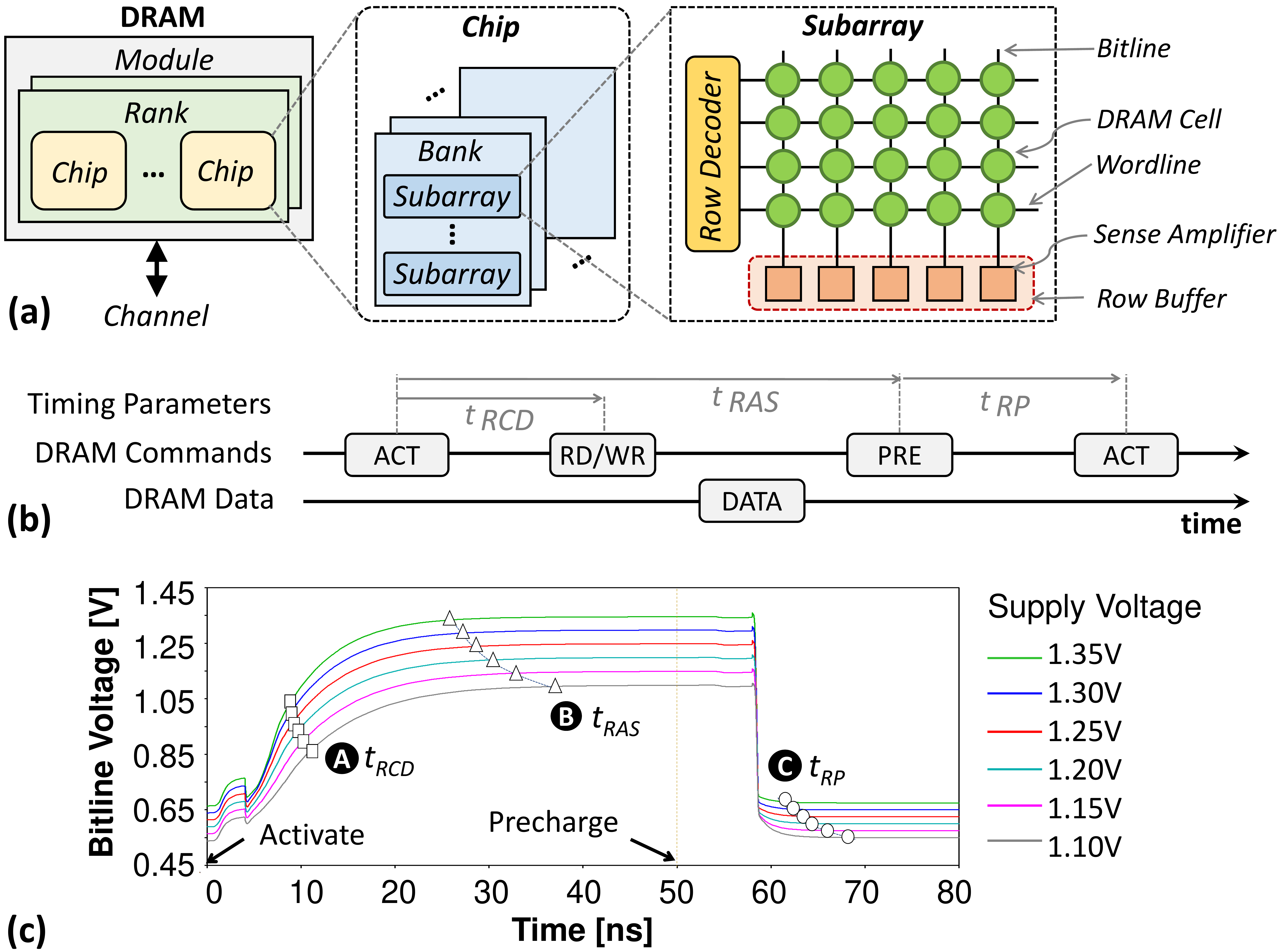}
\caption{(a) The internal organization of commodity DRAM. (b) Diagram of the DRAM commands (i.e., ACT, RW or WR, and PRE) and the DRAM timing parameters (i.e., $t_{RCD}$, $t_{RAS}$, and $t_{RP}$). (c) The dynamics of the DRAM bitline voltage $V_{bitline}$ and the respective timing parameters.}
\label{Fig_DRAMnSpiceDRAM}
\end{figure}

The organization of a DRAM consists of channel, module, rank, chip, bank, subarray, row, and column \citep{Ref_Putra_DRMap_DAC20,Ref_Olgun_PiDRAM_arXiv21}; see Fig.~\ref{Fig_DRAMnSpiceDRAM}(a).
A single DRAM request can access data from multiple DRAM chips within the same rank in parallel.
In each DRAM chip, the request is routed to a specific bank, row, and column address.
When the activation (ACT) command is issued, the requested row is opened and its data are copied to the row buffer.
If the read (RD) command is issued, data in the row buffer can be read. 
If the write (WR) command is issued, data in the row buffer can be replaced with the new one.
In each DRAM request, there are different possible access conditions, i.e., a row buffer hit, miss, and conflict~\citep{Ref_Ghose_WorkloadDRAM_POMACS19}.
A row buffer hit refers to a condition when the requested row is activated and its data are already in the row buffer. 
Hence, the data can be accessed directly without additional operation.
Otherwise, the requested row is still closed, and the condition is either a row buffer miss or conflict. 
A row buffer miss is defined if there is no activated row when a request happens, thus the requested row should be activated before accessing the data. 
Meanwhile, a row buffer conflict is defined when the requested row is still closed, but the row buffer is occupied by another activated row. 
Hence, the activated row should be closed using the precharging (PRE) command, before activating the requested row using the activation (ACT) command.
Fig.~\ref{Fig_DRAMnSpiceDRAM}(b) illustrates the DRAM commands (i.e., ACT, RD or WR, PRE) and their timing parameters (i.e., $t_{RCD}$: row address to column address delay, $t_{RAS}$: row active time, and $t_{RP}$: row precharge time).

%%%%%%%%%%

%%%%%%%%%%
\subsubsection{Reduced-Voltage DRAM}
\label{Sec_Prelim_xDRAMvolt}

We perform extensive experiments using the SPICE simulator and the DRAM circuit model from~\citet{Ref_Chang_Voltron_POMACS17} while considering different supply voltage ($V_{supply}$) values, to characterize the parameters of reduced-voltage DRAM: including the bitline voltage ($V_{bitline}$) and the respective timing parameters (i.e., $t_{RP}$, $t_{RCD}$, and $t_{RAS}$). 
The experimental results are presented in Fig.~\ref{Fig_DRAMnSpiceDRAM}(c), and the obtained parameters are used for further DRAM energy estimation.
The ready-to-access voltage is defined as the condition when $V_{bitline}$ reaches 75\% of $V_{supply}$, which represents the minimum $t_{RCD}$ for reliable DRAM operations, as shown by~\circledB{A}. 
The ready-to-precharge voltage is defined as the condition when $V_{bitline}$ reaches 98\% of $V_{supply}$, which represents the minimum $t_{RAS}$ for reliable DRAM operations, as shown by~\circledB{B}. 
Meanwhile, the ready-to-activate voltage is defined as the condition when the $V_{bitline}$ is within 2\% of $V_{supply}/2$, which represents the minimum $t_{RP}$ for reliable DRAM operations, as shown by~\circledB{C}.

%%%%%%%%%%%%%%%%%%%%%%%%%%%%%%%%%%%%%%%%%%%%%%%%%%%%%%%
%%%%%%%%%%%%%%%%%%%%%%%%%%%%%%%%%%%%%%%%%%%%%%%%%%%%%%%
\section{Error Modeling for Approximate DRAM}
\label{Sec_DRAMerrorModel}

The work of~\citet{Ref_Koppula_EDEN_MICRO19} has proposed four error models that closely fit the real reduced-voltage-based approximate DRAMs as the following. 
\textbf{Error Model-0}: the bit errors follow a \textit{uniform random distribution} across a DRAM bank; 
\textbf{Error Model-1}: the bit errors follow a \textit{vertical distribution} across the bitlines of a DRAM bank;
\textbf{Error Model-2}: the bit errors follow a \textit{horizontal distribution} across the wordlines of a DRAM bank; and
\textbf{Error Model-3}: the bit errors follow a \textit{uniform random distribution} that depends on the content of the DRAM cells.
In this work, we employ the \textbf{DRAM Error Model-0}, 
due to the following reasons: 
(1) it produces errors with high similarity to the real reduced-voltage-based approximate DRAM by using the percentage of weak cells and the error probability in any weak cell; 
(2) it offers a reasonable approximation of other error models, including the approximation of (a) errors across bitlines similar to Error Model-1, (b) errors across wordlines similar to Error Model-2, and (c) uniform random distribution similar to Error Model-3; and 
(3) it provides fast error injection by software. 
Previous work~\citep{Ref_Koppula_EDEN_MICRO19} also employed the DRAM Error Model-0 majorly due to similar reasons.

%%%%%%%%%%%%%%%%%%%%%%%%%%%%%%%%%%%%%%%%%%%%%%%%%%%%%%%
%%%%%%%%%%%%%%%%%%%%%%%%%%%%%%%%%%%%%%%%%%%%%%%%%%%%%%%
\section{EnforceSNN Framework}
\label{Sec_EnforceSNN}

%%%%%%%%%%%%%%%%%%%%%%%
\subsection{Overview}
\label{Sec_EnforceSNN_Overview}

\begin{figure}[t]
\centering
\includegraphics[width=\linewidth]{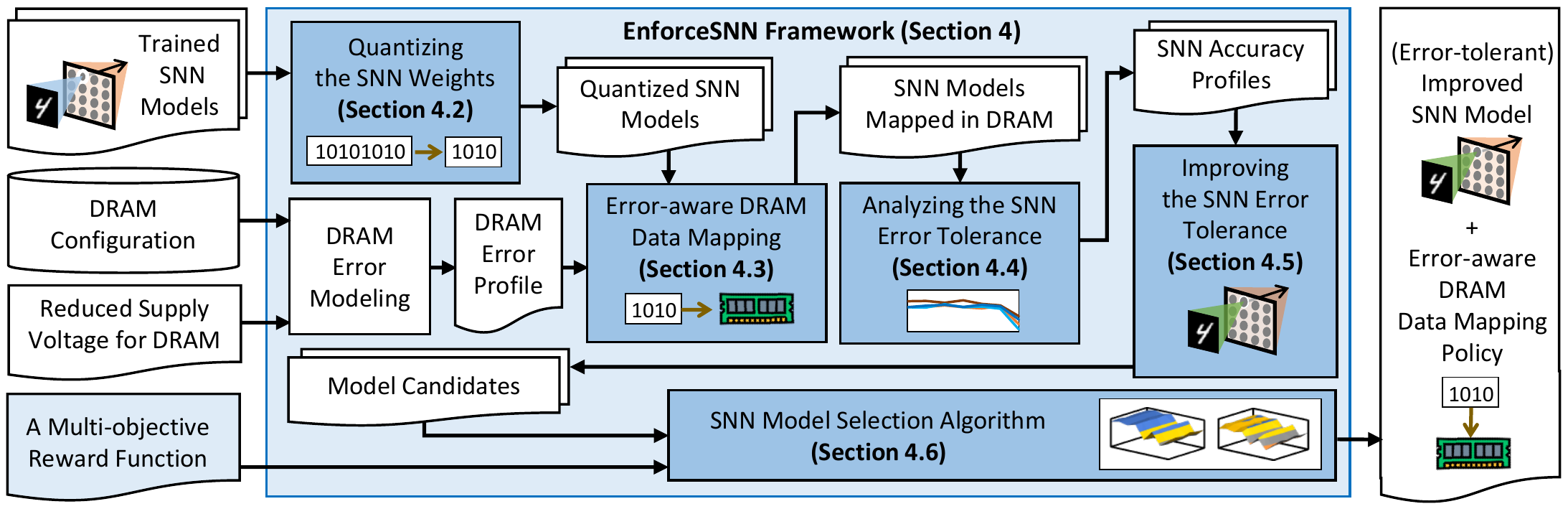}
\caption{The EnforceSNN framework and its novel mechanisms (highlighted in blue boxes).}
\label{Fig_EnforceSNN}
\end{figure}

Our EnforceSNN framework employs several key steps as shown in Fig.~\ref{Fig_EnforceSNN}.
First, we \textit{quantize the SNN weights} to reduce memory footprint and DRAM access energy 
(Section~\ref{Sec_EnforceSNN_QuantWGH}). 
Second, we devise \textit{an error-aware DRAM data mapping policy} to optimize the DRAM energy-per-access (Section~\ref{Sec_EnforceSNN_DRAMmap}).
These optimizations contribute to 4.1x inference speed-up and 84.9\% DRAM access energy saving (Section~\ref{Sec_Res}). 
Then, we \textit{analyze the SNN error tolerance} to understand the accuracy profile of SNN inference under different BER values (Section~\ref{Sec_EnforceSNN_AnalyzeSNN}).
We leverage this information to \textit{develop an efficient FAT technique that improves the SNN error tolerance} (Section~\ref{Sec_EnforceSNN_ImproveSNN}).
These fault tolerance techniques contribute to 2.71x retraining speed-up without accuracy loss for $BER \leq 10^{-3}$ (Section~\ref{Sec_Res}). 
We also \textit{develop an SNN model selection algorithm} to find a model that provides good trade-offs among accuracy, memory, and energy consumption (Section~\ref{Sec_EnforceSNN_Algo}).
Details of these mechanisms are explained in the following subsections.

%%%%%%%%%%%%%%%%%%%%%%%
\subsection{Quantizing the SNN Weights}
\label{Sec_EnforceSNN_QuantWGH}

\begin{figure}[t]
    \centering
    \includegraphics[width=0.75\linewidth]{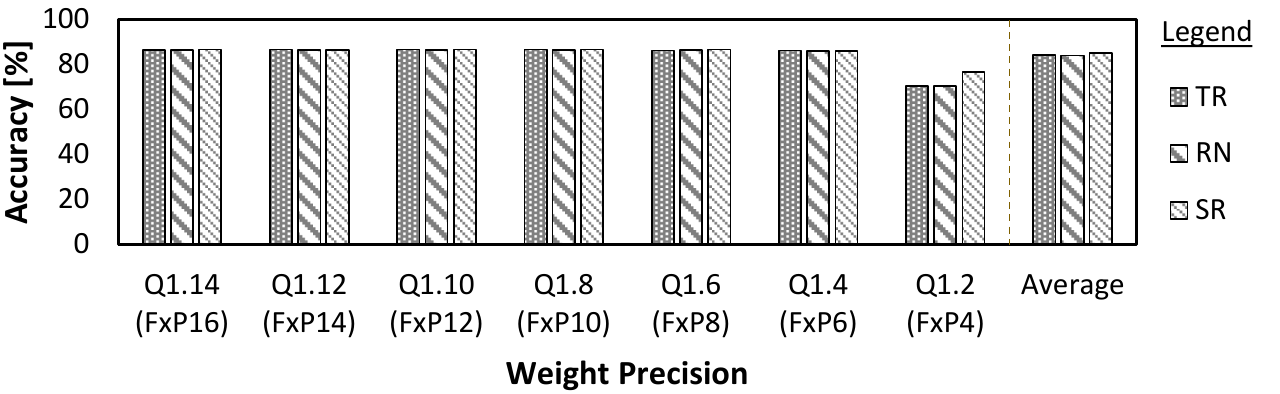}
    \caption{The accuracy of a 900-neuron network on the MNIST across different precision levels and rounding schemes, i.e., truncation (TR), round to the nearest (RN), and stochastic rounding (SR). 
    Here, we follow the definition of TR, RN, and SR from the work of~\cite{Ref_Hopkins_Rounding_RSTA20}. 
    Note, the fixed-point format $Qi.f$ can also be represented as FxP(1+$i$+$f$).
    These results show that employing TR on top of the FxP8 precision leads to competitive accuracy as compared to other rounding schemes (i.e., RN and SR).}
    \label{Fig_ObserveRounding}
\end{figure}

\textit{We perform weight quantization to substantially reduce the memory footprint and the number of DRAM accesses, which lead to DRAM energy saving}.
The reason is that quantization is a prominent technique for reducing the memory footprint of neural networks without decreasing the accuracy significantly~\citep{Ref_Gupta_DLPrecision_ICML15,Ref_Micikevicius_MixedPrecision_ICLR18}.
Moreover, it is the first effort to study and exploit SNN weight quantization considering approximation errors in DRAM, thereby providing new insights as compared to previous studies on SNN weight quantization. 
Our weight quantization considers the fixed-point format which can be represented as $Qi.f$. 
It denotes 1 sign bit, $i$ integer bits, and $f$ fractional bits, and follows the 2’s complement format (i.e., signed $Qi.f$). 
Here, the range of representable values is $[\minus 2^{i},2^{i}-2^{\minus f}]$ with the precision of $\epsilon = 2^{\minus f}$.
We select the ``signed $Qi.f$" format to show that our EnforceSNN framework provides a generic solution with high applicability for different variants of bio-plausible learning rules (e.g., STDP variants)  which may lead to positive or negative weight values~\citep{Ref_Azghadi_Plasticity_JPROC14, Ref_Diehl_STDPmnist_FNCOM15}.
To do this, we perform a fixed-point quantization to the trained SNN weights using a specific rounding scheme, such as truncation, round to the nearest, or stochastic rounding.
For a study case, we select the truncation as it provides competitive accuracy with low computational complexity~\citep{Ref_Putra_QSpiNN_IJCNN21,Ref_Putra_lpSpikeCon_arXiv22,Ref_Putra_tinySNN_arXiv22}.
To illustrate this, we evaluate the impact of different rounding schemes on the accuracy through an experimental case study, and the results are shown in Fig.~\ref{Fig_ObserveRounding}.
Truncation (TR) keeps the $f$ bits and removes the other fractional bits.
Therefore, the output fixed-point for the given real number $x$ with $Qi.f$ format is defined as $TR(x, Qi.f) = \lfloor x \rfloor$.
In our SNN model, we employ the pair-based weight-dependent STDP learning rule from the work of~\citet{Ref_Diehl_STDPmnist_FNCOM15} that bounds each weight value ($w$) within the defined range, i.e., $w = [0,1]$. 
Consequently, applying the truncation to the weights will round the value down.
In this work, we consider 8-bit fixed-point with ``signed $Q1.6$" and 2's complement format (i.e., 1 sign bit, 1 integer bit and 6 fractional bits), since it provides high accuracy for SNNs under unsupervised learning scenarios~\citep{Ref_Putra_QSpiNN_IJCNN21}.
Note, we can also employ the ``unsigned $Qi.f$" format without sign bit to represent the 8-bit non-negative weights (i.e., 1 integer bit and 7 fractional bits) in the EnforceSNN if desired.
For both ``signed $Qi.f$" and ``unsigned $Qi.f$" formats, 1 bit for integer part is required for representing the maximum possible weight value (i.e., $w=1$).

\textbf{Quantization Steps:} 
We quantize only the weights through the simulated quantization approach, which represents the weight values under fixed-point format, and performing computations under floating-point format~\citep{Ref_Krishnamoorthi_Whitepaper_arXiv18, Ref_Jacob_QuantNN_CVPR18, Ref_Gholami_QuantSurvey_arXiv21, Ref_vanBaalen_SimulatedQuant_ICCV22}. 
To perform quantization, we convert the weight values from 32-bit floating-point format (FP32) to 8-bit fixed-point format (signed $Q1.6$) by constructing their 8-bit binary representations under 32-bit integer format (INT32), thereby conveniently performing bit-wise modification and rounding operation while considering the sign and the rounding scheme (i.e., truncation).
Afterwards, we convert the quantized weight values (INT32) to FP32 format through casting and then normalizing the values by $2^f$.
Hence, the 8-bit binary representations of quantized weight values are saved in FP32 and can be used in the FP32-based arithmetic computations.

\textbf{DRAM Error Injection:}
\textit{If there is no DRAM error}, the quantization steps are performed and the quantized weight values (in FP32) can be used for computations in SNN processing.
\textit{If DRAM errors exist}, the quantization steps are performed while considering the DRAM error injection. 
These errors are injected to the 8-bit binary representations of quantized weights (in INT32) under a specific DRAM data mapping policy.
Afterwards, we convert the binary representations of quantized weights (in INT32) to FP32 format, so that the quantized weight values can be used for computations in SNN processing.
    
%%%%%%%%%%%%%%%%%%%%%%%
\subsection{Error-aware DRAM Data Mapping Policy}
\label{Sec_EnforceSNN_DRAMmap}

\begin{figure}[t]
\centering
\includegraphics[width=0.8\linewidth]{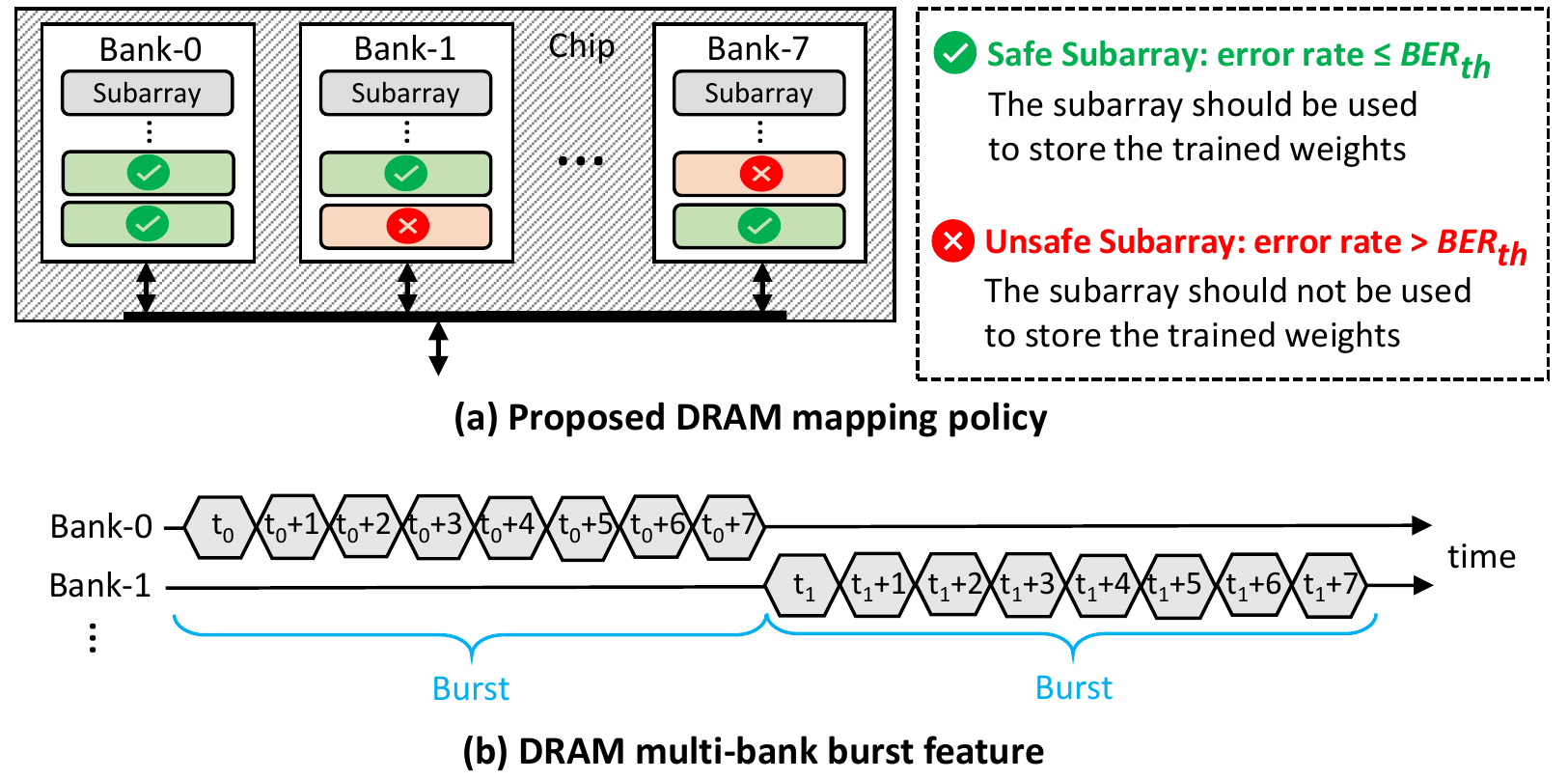}
\vspace{-0.2cm}
\caption{(a) Our proposed DRAM mapping policy, leveraging subarray-level granularity. (b) The timing diagram of the DRAM multi-bank burst feature.}
\label{Fig_DRAMmap_burst}
\end{figure}

It is important to map the SNN model properly in DRAM to ensure that (1) the weights are minimally affected by errors in DRAM so that the accuracy is maintained, and (2) the DRAM energy-per-access is optimized.
Towards this, \textit{we devise and employ an error-aware DRAM mapping policy to place the SNN weights in DRAM, while optimizing the DRAM
energy-per-access}. 
The proposed DRAM mapping policy is illustrated in Fig.~\ref{Fig_DRAMmap_burst}(a), and its key ideas are explained in the following.

\begin{figure}[t]
\centering
\includegraphics[width=\linewidth]{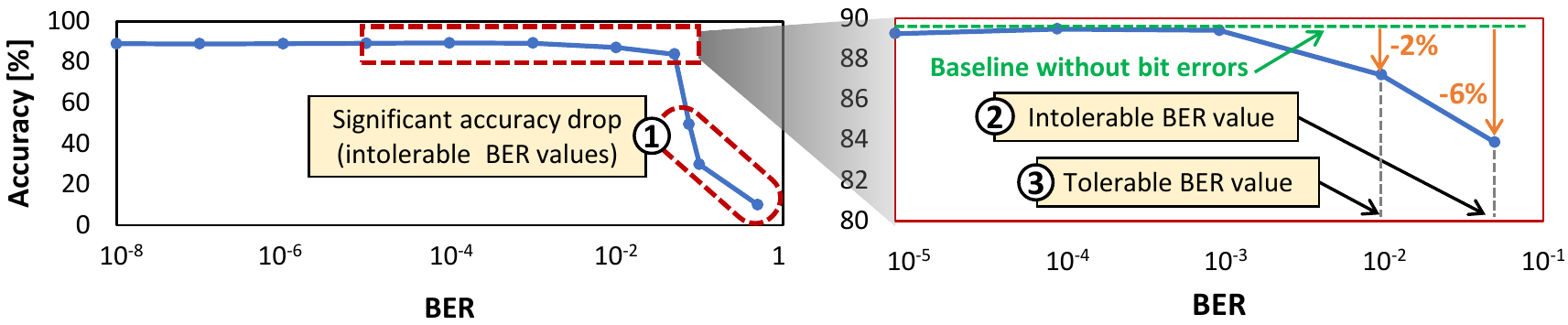}
\vspace{-0.5cm}
\caption{The test accuracy profile of a 900-neuron network across different BER values, showing the tolerable and intolerable BER.
This network has a fully-connected architecture like in Fig.~\ref{Fig_SNN_n_LIF}(a) where each input pixel is connected to all neurons, and the output of each neuron is connected to other neurons for performing inhibition, thereby providing competition among neurons to correctly recognize the input class.}
\label{Fig_BERth}
\vspace{-0.5cm}
\end{figure}

\begin{enumerate}
    \item The weights are mapped in the appropriate DRAM part (e.g., chip, bank, or subarray), whose error rate meets the BER requirement, i.e., $\leq$ the maximum tolerable BER ($BER_{th}$). 
    Here, we consider the subarray-level granularity for data mapping, since it allows us to exploit the following features.
    \begin{itemize}
        \item \textit{The multi-bank burst feature}, which is available in commodity DRAM, can be employed to increase the throughput. 
        Its timing diagram is illustrated in Fig.~\ref{Fig_DRAMmap_burst}(b).
        \item \textit{The subarray-level parallelism}, which is available in novel DRAM architectures~\citep{Ref_Kim_SALP_ISCA12}, can also be employed to increase the throughput.
    \end{itemize}
    We determine $BER_{th}$ through experiments that investigate the accuracy profile of a network across different BER values.
    Fig.~\ref{Fig_BERth} shows the experimental results for a 900-neuron network. 
    If the accuracy scores are significantly lower than the baseline accuracy without bit errors (i.e., $>$3\% accuracy degradation), we refer the respective BER values as \textit{intolerable BER}, as shown by \circled{1} and \circled{2} in Fig.~\ref{Fig_BERth}.
    Otherwise, we define them as \textit{tolerable BER}.
    For instance, \circled{3} in Fig.~\ref{Fig_BERth} shows an accuracy that is associated with a tolerable BER. 
    Based on this discussion, we define $BER_{th}=10^{-2}$. 
    \item The weights are mapped in a way to maximize the row buffer hits for optimizing the DRAM energy-per-access while exploiting the multi-bank burst feature for maximizing the data throughput. 
    The reason is that a row buffer hit incurs the lowest DRAM access energy than other access conditions (i.e., a row buffer miss or conflict), as suggested by our experimental results in Fig.~\ref{Fig_ObserveApproxDRAM}(b). 
\end{enumerate}

To efficiently implement the above ideas, we devise Alg.~\ref{Alg_DRAMmapping} with the following key steps. 
First, we identify the subarrays whose error rate $\leq$ $BER_{th}$ and refer them as \textit{the safe subarrays}, which should be used for storing the weights. 
Otherwise, we refer the subarrays whose error rate $>$ $BER_{th}$ as \textit{the unsafe subarrays}, which should not be used for storing the weights. 
This step is represented in line 7 of Alg.~\ref{Alg_DRAMmapping}.
Second, to maximize the row buffer hits and exploit the multi-bank burst feature, the data mapping in each DRAM chip should follow the following policy (represented in lines 3-8 of Alg.~\ref{Alg_DRAMmapping}).

\begin{itemize}
    \item \textbf{Step-1:} 
    Assume that we consider mapping data in a target subarray of the target bank with the following initial indices, i.e., \textit{subarray\_index = i, bank\_index = j}. 
    \item \textbf{Step-2:} 
    If the target subarray is a safe subarray, then we prioritize mapping the data in different columns of the same row for maximizing row buffer hits. 
    Otherwise, this subarray is not utilized and we move to another target subarray in a different bank (\textit{subarray\_index = i, bank\_index += 1}). 
    Then, we perform \textbf{Step-2} again. 
    If all columns in the same row across all banks are filled or unavailable, then we move to another subarray in the initial bank (\textit{subarray\_index += 1, bank\_index = j}) to exploit subarray-level parallelism, if applicable.
    \item \textbf{Step-3:} 
    In the target subarray, the remaining data are mapped in the same fashion as \textbf{Step-2}. 
    When all columns in the same row of all safe subarrays across all banks are filled, then the remaining data are placed in a different row of the initial target subarray and bank (\textit{subarray\_index = i, bank\_index = j}). 
    Afterwards, we perform \textbf{Step-2} to \textbf{Step-3} again until all data are mapped in a DRAM chip.
    If some data remain but there are no available spaces in a DRAM chip, then we move to \textbf{Step-4}. 
    \item \textbf{Step-4:} 
    The remaining data are mapped using \textbf{Step-1} to \textbf{Step-3} in different DRAM chips, ranks, and channels, respectively if applicable.
\end{itemize}

%%%%%%%%%%%%%%%%%%%%%%%
\subsection{Analyzing the SNN Error Tolerance}
\label{Sec_EnforceSNN_AnalyzeSNN}
\vspace{-0.2cm}

\begin{figure}[t]
\centering
\includegraphics[width=0.86\linewidth]{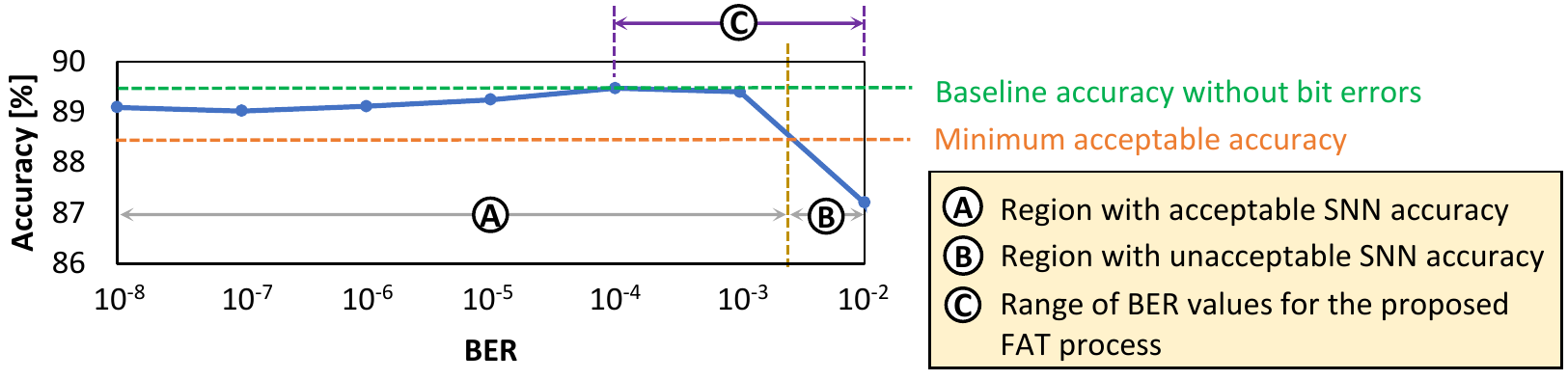}
\vspace{-0.2cm}
\caption{The test accuracy profile of a 900-neuron network shows the region with acceptable accuracy, the region with unacceptable accuracy, and the range of BER values for the proposed fault-aware training. 
Note, this network is the same with the one in Fig.~\ref{Fig_BERth}.
Here, the observation focuses on the BER values that can be considered in the retraining process, thereby having a smaller range than the one in Fig.~\ref{Fig_BERth}.}
\label{Fig_AnalysisBER}
\vspace{-0.2cm}
\end{figure}

Previous discussion highlights that bit errors in the SNN weights can degrade the accuracy, as they change the weight values and deviate the neuron behavior from the correct classification.
Therefore, SNN error tolerance should be improved, so that the SNN model can achieve high accuracy even in the presence of high error rate.
To effectively enhance the SNN error tolerance, it is important to understand the SNN accuracy profile under DRAM errors.
Towards this, \textit{our EnforceSNN analyzes the accuracy profile of the SNN model considering the data mapping pattern in DRAM and different BER values}. 
We observe that the accuracy profile typically has acceptable accuracy (i.e., within 1\% accuracy degradation from the baseline without errors) when BER is low, and has notable accuracy degradation when BER is high. 
Therefore, we classify the accuracy profile into two regions: \circled{A} a region with acceptable accuracy, and \circled{B} a region with unacceptable accuracy, as shown in Fig.~\ref{Fig_AnalysisBER}.
These insights will be leveraged for developing an efficient enhancement technique for improving the SNN error tolerance in Section~\ref{Sec_EnforceSNN_ImproveSNN}.

%%% --------
\begin{algorithm}[t]
\scriptsize
\caption{\color{black} The proposed DRAM mapping policy}
\label{Alg_DRAMmapping}
\begin{algorithmic}[1]
\renewcommand{\algorithmicrequire}{\textbf{INPUT:}}
\renewcommand{\algorithmicensure}{\textbf{OUTPUT:}}
\REQUIRE \textbf{(1)} DRAM ($DRAM$): number of channel ($n_{ch}$), number of rank-per-channel ($n_{ra}$), number of chip-per-rank ($n_{cp}$), number of bank-per-chip ($n_{ba}$), number of subarray-per-bank ($n_{su}$), number of row-per-subarray ($n_{ro}$), number of column-per-row ($n_{co}$);\\
\textbf{(2)} Bit error rate (BER): BER of a subarray ($BER\_{subarray}$), maximum tolerable BER ($BER_{th}$);\\
\textbf{(3)} Data ($data$); \\
\ENSURE DRAM ($DRAM$); \\
\vspace{0.1cm}
\renewcommand{\algorithmicrequire}{\textbf{BEGIN}}
\renewcommand{\algorithmicensure}{\textbf{END}}
\REQUIRE \hspace{0.1cm} \\   
    \textbf{Process}: \\
    \FOR{$ch = 0$ to $(n_{ch}-1)$}
    \FOR{$ra = 0$ to $(n_{ra}-1)$}
    \FOR{$cp = 0$ to $(n_{cp}-1)$}
    \FOR{$ro = 0$ to $(n_{ro}-1)$}
    \FOR{$su = 0$ to $(n_{su}-1)$}
    \FOR{$ba = 0$ to $(n_{ba}-1)$}
      \IF{$BER\_subarray[ch, ra, cp, ba, su] \leq BER_{th}$}
        \FOR{$co = 0$ to $(n_{co}-1)$} 
          \STATE $DRAM[ch, ra, cp, ba, su, ro, co] \leftarrow  data$;
        \ENDFOR
      \ENDIF
    \ENDFOR
    \ENDFOR
    \ENDFOR
    \ENDFOR
    \ENDFOR
    \ENDFOR  
    \RETURN $DRAM$;
\ENSURE 
\end{algorithmic}
\end{algorithm}
\setlength{\textfloatsep}{12pt}
%%% --------

%%%%%%%%%%%%%%%%%%%%%%%
\subsection{Improving the SNN Error Tolerance}
\label{Sec_EnforceSNN_ImproveSNN}
\vspace{-0.2cm}

\textit{Our EnforceSNN enhances the SNN error tolerance through the fault-aware training (FAT) technique that incorporates the error profile of the approximate DRAM}.
We consider efficiently performing FAT for minimizing training time, energy consumption, and carbon emission during the retraining process~\citep{Ref_Strubell_Carbon_ACL19,Ref_Strubell_Carbon_AAAI20}, by conducting a small yet effective number of iterations for the retraining process, while avoiding \textit{accuracy collapse}.
Accuracy collapse is defined as a significant accuracy degradation due to training divergence that is caused by introducing high BER immediately at the beginning of the retraining process~\citep{Ref_Koppula_EDEN_MICRO19}.
The proposed FAT technique has the following key steps, which are also presented in Alg.~\ref{Alg_OurFAT}.
\vspace{-0.2cm}
\begin{itemize}
    \vspace{-0.2cm}
    \item \textbf{Step-1:}
    We define the range of BER values that will be incorporated in the training process to make the SNN model adaptable to DRAM errors, as shown by region-\circled{C} in Fig.~\ref{Fig_AnalysisBER}. 
    We incorporate (1) BER values in region-\circled{A} that are close to region-\circled{B}, and (2) all BER values in region-\circled{B}, in the training process.
    Specifically, we consider the two highest BER values in region-\circled{A} in the training process to make the model adapt to high fault rates safely, without suffering from significant accuracy degradation (i.e., accuracy collapse) with less training time. 
    \item \textbf{Step-2:}
    The bit errors in DRAM are generated for different BER values (which correspond to different $V_{supply}$ values), based on the DRAM error model-0 that follows a uniform random distribution across a DRAM bank.
    \item \textbf{Step-3:} 
    The generated bit errors are then injected into the DRAM cell locations, and the weight bits in these locations are flipped. 
    In this step, we consider the proposed DRAM data mapping discussed in Section~\ref{Sec_EnforceSNN_DRAMmap} for maximizing the row buffer hits and exploiting the multi-bank burst feature. 
    \item \textbf{Step-4:} 
    Afterwards, we include the generated bit errors in the retraining by incrementally increasing the BER from the minimum rate to a maximum one following the defined range of BER values from \textbf{Step-1}. 
    We increase the BER value after each epoch of retraining by a defined ratio (e.g., 10x of the previous error rate). 
    In this manner, the SNN model is gradually trained to tolerate DRAM errors from the defined lowest rate to the maximum one, thereby carefully improving the SNN error tolerance.
\end{itemize}

%%%%%%%%%%%%%%%%%%%%%%%
\subsection{Algorithm for SNN Model Selection}
\label{Sec_EnforceSNN_Algo}

From the previous steps, we may get different sizes of error-tolerant SNN models as potential solutions for the given embedded applications.
Therefore, we need to consider design trade-offs to select the most appropriate model for the given accuracy, memory, and energy constraints.
Towards this, \textit{we propose an algorithm to quantify the trade-off benefits of the SNN model candidates using our proposed reward function, and then select the one with the highest benefit.}  
The idea of our multi-objective reward function ($R$) is to prioritize the model that has high accuracy, small memory, and low energy consumption. 
The reward $R$ is defined as the resultant between the accuracy with the memory and energy consumption, as expressed in Eq.~\ref{Eq_Reward}.
In this equation, $acc_x$ denotes the accuracy of the investigated SNN model ($x$).
$m_{norm}$ denotes the normalized memory, which is defined as the ratio between the memory footprint of the investigated model ($mem_x$) and the floating-point model ($mem_{fp}$); see Eq.~\ref{Eq_MemoryNorm}.
The memory footprint of the model is estimated by leveraging the number of weights ($N_{wgh}$) and the corresponding bit-width ($BW_{wgh}$); see Eq.~\ref{Eq_Memory}.
Meanwhile, $E_{norm}$ denotes the normalized energy consumption, which is defined as the ratio between the DRAM access energy of the approximate DRAM ($E_{DRAM\_approx}$) and the accurate one ($E_{DRAM\_accurate}$) for the investigated model; see Eq.~\ref{Eq_EnergyNorm}.
To define the significance of memory and energy consumption with respect to the accuracy when calculating $R$, we employ $\mu$ and $\varepsilon$ as the adjustable trade-off variables for memory and energy consumption, respectively. 
Here, $\mu$ and $\varepsilon$ are the non-negative real numbers.
%
%%% --------
\begin{algorithm}[t]
\scriptsize
\caption{The proposed FAT technique}
\label{Alg_OurFAT}
\begin{algorithmic}[1]
\renewcommand{\algorithmicrequire}{\textbf{INPUT:}}
\renewcommand{\algorithmicensure}{\textbf{OUTPUT:}}
\REQUIRE \textbf{(1)} Baseline pre-trained SNN: model ($model_0$), accuracy ($model_0.acc$); \\
\textbf{(2)} DRAM error model ($DRAMerr$); \\
\textbf{(3)} BER for retraining: error rates ($BER$), number of error rates ($N_{BER}$); \\
\textbf{(4)} Training dataset: samples ($S_{train}$), number of samples ($N_{train}$); \\
\textbf{(5)} Test dataset: samples ($S_{test}$), number of samples ($N_{test}$); \\
\vspace{0.1cm}
\ENSURE \textbf{(1)} Improved SNN: model ($model_1$), accuracy ($model_1.acc$); \\ 
\vspace{0.1cm}
\renewcommand{\algorithmicrequire}{\textbf{BEGIN}}
\renewcommand{\algorithmicensure}{\textbf{END}}
\REQUIRE \hspace{0.1cm} \\   
    \textbf{Initialization}: \\
     \STATE $model_{temp} = model_0$; \\
     \STATE $model_1 = model_0$; \\
     \STATE $model_1.acc = 0$; \\
    \textbf{Process}: \\
      \FOR{$i = 0$ to $(N_{BER}-1)$}
        \STATE $error\_map = DRAMerr(BER[i])$; // error generation 
        \STATE inject $error\_map$ into $model_{temp}$; // error injection 
        \FOR{$r = 0$ to $(N_{train}-1)$}
          \STATE train $model_{temp}$ with $S_{train}[r]$; // train
        \ENDFOR
        \FOR{$s = 0$ to $(N_{test}-1)$}
          \STATE test $model_{temp}$ with $S_{test}[s]$; // test
        \ENDFOR
        \IF{$model_{temp}.acc > model_1.acc$}
          \STATE $model_1 = model_{temp}$;
          \STATE $model_1.acc = model_{temp}.acc$;
        \ENDIF
      \ENDFOR
    \RETURN $model_1$;
\ENSURE 
\end{algorithmic}
\end{algorithm}
%%% --------
%
\begin{equation}
\small
R(acc_x,m_{norm},E_{norm}) = acc_x - (\mu \cdot m_{norm} + \varepsilon \cdot E_{norm})
\label{Eq_Reward}
\end{equation}
\begin{equation}
\small
m_{norm} = \frac{mem_x}{mem_{fp}}  
\label{Eq_MemoryNorm}
\end{equation}
\begin{equation}
\small
mem = N_{wgh} \cdot BW_{wgh}  
\label{Eq_Memory}
\end{equation}
\begin{equation}
\small
\begin{split}
E_{norm} = \frac{E_{DRAM\_approx}}{E_{DRAM\_accurate}}  
\end{split}
\label{Eq_EnergyNorm}
\end{equation}

%%%%%%%%%%%%%%%%%%%%%%%%%%%%%%%%%%%%%%%%%%%%%%%%%%%%%%%
%%%%%%%%%%%%%%%%%%%%%%%%%%%%%%%%%%%%%%%%%%%%%%%%%%%%%%%
\section{Evaluation Methodology}
\label{Sec_Eval}

\begin{figure}[t]
\centering
\includegraphics[width=0.82\linewidth]{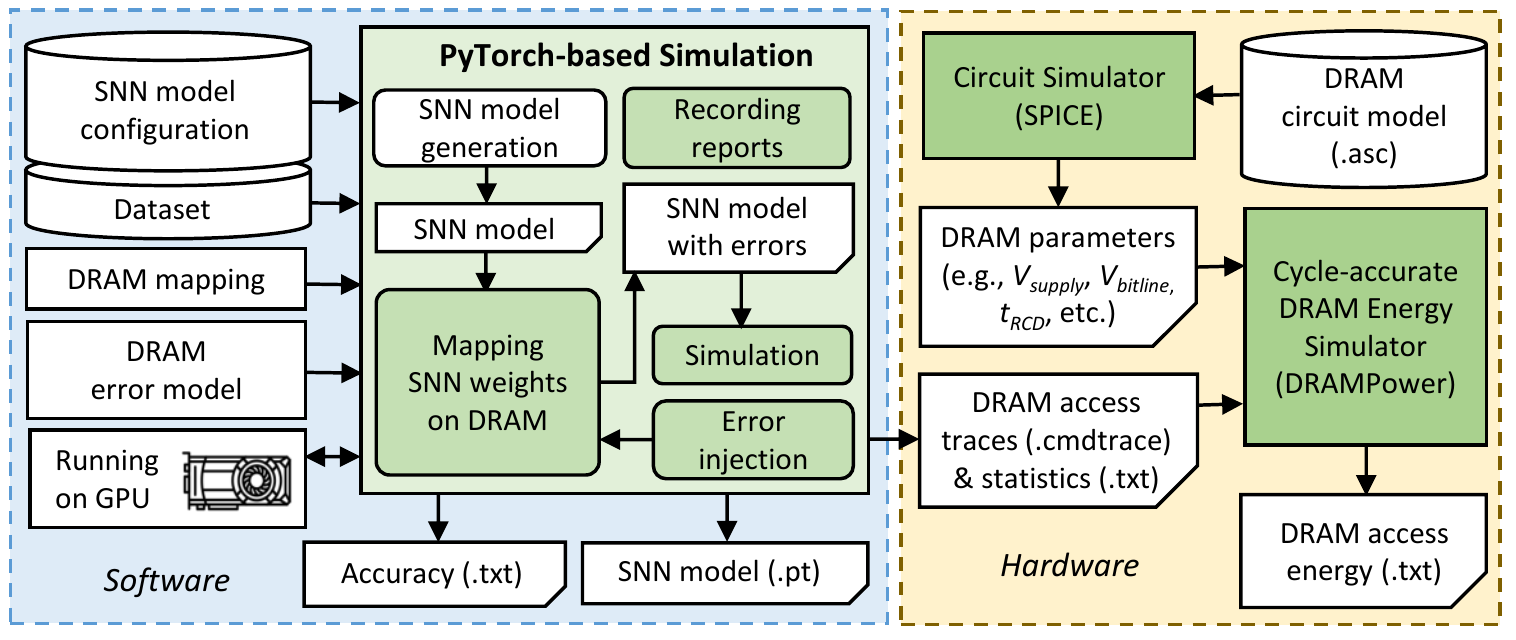}
\caption{Our experimental setup and tools flow.}
\label{Fig_ExpSetup}
\end{figure}

Fig.~\ref{Fig_ExpSetup} shows the experimental setup and tools flow for evaluating our EnforceSNN framework, which are explained in the following.

\textbf{Accuracy Evaluation:} 
We employ PyTorch-based simulations~\citep{Ref_Hazan_BindsNET_FNINF18} with 32-bit floating-point (FP32) and 8-bit fixed-point precision (i.e., FxP8 with ``signed $Q1.6$" and ``unsigned $Q1.7$" formats) that run on a multi-GPU machine, i.e., Nvidia GeForce RTX 2080 Ti.
For network architecture, we consider the fully-connected network with a different number of excitatory neurons, which are referred to as N-\textit{i} for conciseness with \textit{i} denoting the number of excitatory neurons.
We use the rate coding for converting each input pixel into a spike train, and use the MNIST and Fashion MNIST datasets. 
For comparison partners, we use the SNN model which is pre-trained without considering DRAM errors as the baseline.
We perform an epoch of STDP-based unsupervised learning through 60K experiments for each retraining process considering each combination of the SNN model, dataset, and training BER. 
Afterwards, we perform inference through 10K experiments for each combination of the SNN model, dataset, and testing BER. 

\textbf{DRAM Error Generation and Injection:} 
First, we generate bit errors based on the DRAM error model-0, and inject them into the DRAM cell locations, while considering the data mapping policy in DRAM. 
Afterwards, the weight bits that are stored in the DRAM cell locations with errors, will be flipped. 
For the baseline data mapping, we place the weight bits in the subsequent address in a DRAM bank to maximize the DRAM burst feature, and if a DRAM bank is filled, then the weight bits are mapped in a different bank of the same DRAM chip.
Meanwhile, we use the proposed DRAM mapping in Alg.~\ref{Alg_DRAMmapping} for our EnforceSNN.

\textbf{DRAM Energy Evaluation:} 
We use the DRAM circuit model from the work of~\citep{Ref_Chang_Voltron_POMACS17} and the SPICE simulator to extract the DRAM operational parameters (e.g., $V_{supply}$, $V_{bitline}$, $t_{RCD}$, $t_{RAS}$, $t_{RP}$), while considering the configuration of LPDDR3-1600 4Gb DRAM which is representative for the main memory of embedded systems. 
The accurate DRAM operates with 1.35V of $V_{supply}$, while the approximate one operates with 1.025V-1.325V of $V_{supply}$. 
Afterwards, we use the state-of-the-art cycle-accurate DRAMPower~\citep{Ref_Chandrasekar_DRAMPower} that incorporates the DRAM access traces and statistics as well as the extracted DRAM parameters for estimating the DRAM access energy.

%%%%%%%%%%%%%%%%%%%%%%%%%%%%%%%%%%%%%%%%%%%%%%%%%%%%%%%
%%%%%%%%%%%%%%%%%%%%%%%%%%%%%%%%%%%%%%%%%%%%%%%%%%%%%%%
\section{Results and Discussion}
\label{Sec_Res}

%%%%%%%%%%%%%%%%%%%%%%%
\subsection{Improvements of the SNN Error Tolerance}
\label{Sec_Res_Accuracy}

\begin{figure}[t]
\centering
\includegraphics[width=\linewidth]{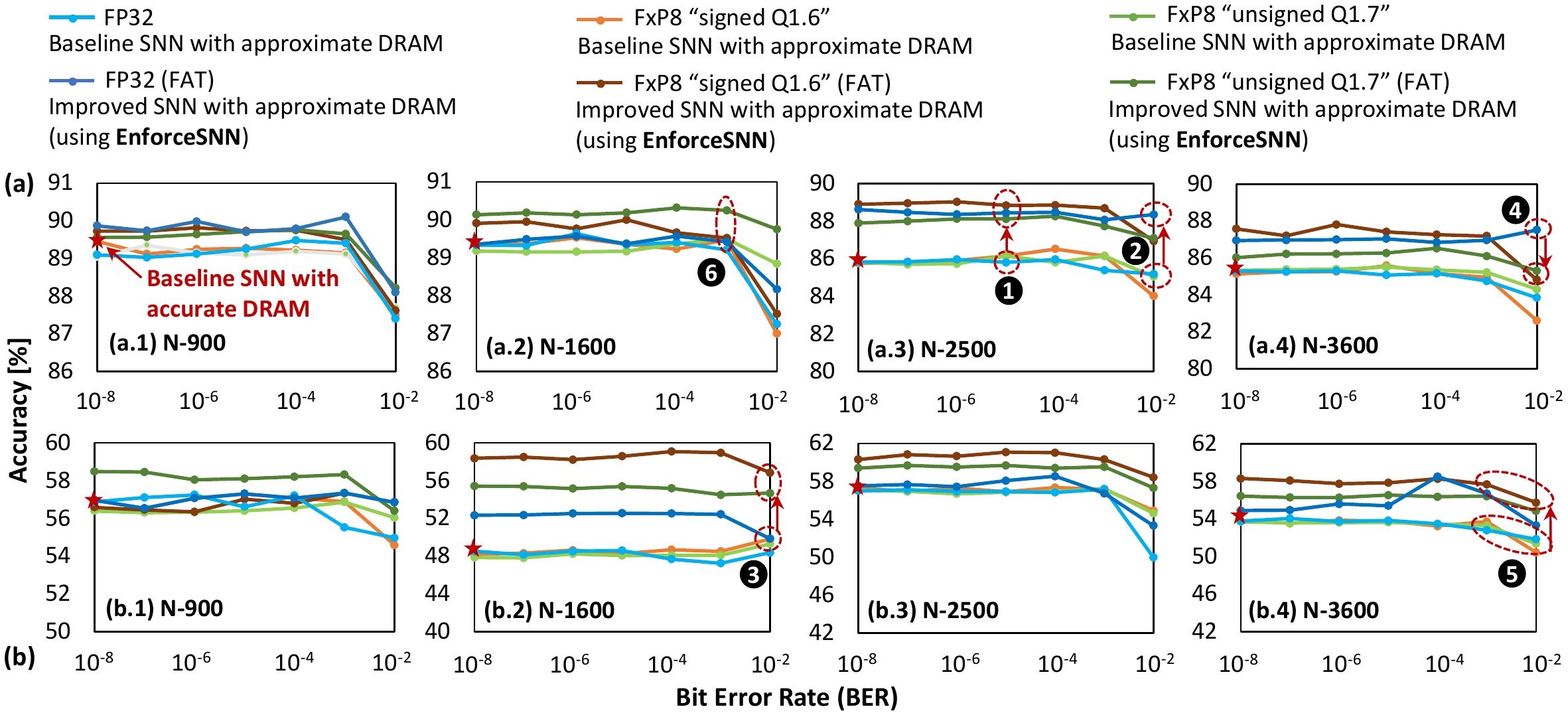}
\caption{The accuracy of the baseline model with accurate and approximate DRAM, as well as the EnforceSNN-improved model with approximate DRAM for (a) MNIST and (b) Fashion MNIST datasets, across different precision levels, different BER values, and different network sizes.}
\label{Fig_Results_Accuracy}
\end{figure}

Fig.~\ref{Fig_Results_Accuracy} shows the accuracy of the baseline model and the EnforceSNN-improved model with accurate and approximate DRAM across different BER values, precision levels, i.e., FP32 and FxP8 (``signed $Q1.6$" and ``unsigned $Q1.7$"), network sizes, and workloads (i.e., the MNIST and Fashion MNIST datasets).
In general, we observe that the baseline model with approximate DRAM achieves lower accuracy than the baseline model with accurate DRAM, and the accuracy decreases as the BER increases.
These trends are observed across different weight precision levels, network sizes, and datasets.
The reason is that, the weights are changed (i.e., flipped) if they are stored in the faulty DRAM cells, and these weights are not trained to adapt to such bit flips. 
Therefore, the corresponding neuron behavior deteriorates from the expected behavior, hence decreasing accuracy. 
On the other hand, the EnforceSNN-improved model with approximate DRAM improves accuracy over the baseline model with accurate and approximate DRAM, across different BER values, network sizes, and datasets, as shown in~\circledB{1}.
We also observe that, the EnforceSNN-improved model with approximate DRAM improves the accuracy over the baseline model with accurate and approximate DRAM, even in the high error rate case (i.e., $BER=10^{-2}$), as shown in~\circledB{2} for FP32 and~\circledB{3} for FxP8 weight precision levels.
The reason is that, our EnforceSNN incorporates the error profiles from the approximate DRAM across different BER values in the training process, which makes the SNN model adaptive to the presence of DRAM errors, thereby improving the SNN error tolerance.
For the MNIST dataset, a high error rate (i.e., $BER=10^{-2}$) typically decreases the accuracy of the SNN-FxP8 more than the SNN-FP32, as shown in~\circledB{4}. 
The reason is that, the MNIST dataset has a narrow weight distribution in each class to represent its digit features, hence bit errors may change the weight values significantly in the FxP8 precision than the FP32 due to its shorter bit-width. 
As a result, the corresponding neuron behavior deteriorates from its ideal behavior, hence degrading accuracy.
For the Fashion MNIST dataset, the SNN-FxP8 may achieve higher accuracy than the SNN-FP32 in some cases, as shown in~\circledB{5}. 
The potential reason is the following.
The Fashion MNIST dataset has relatively more complex features than the MNIST dataset, hence having a wider weight distribution in each class to represent its various features which may overlap with features from other classes (i.e., non-unique features).
Then, the quantization removes these non-unique features by eliminating the less significant bits of the trained weights (i.e., like the denoising effect), and the retraining makes the quantized weights adaptive to bit flips, thereby leading to higher accuracy than the non-quantized ones.
Furthermore, we also observe that the accuracy of the SNN-FxP8 starts showing notable degradation at a high error rate (i.e., $BER=10^{-2}$).
For quantized models, in general, the ``unsigned $Q1.7$" and ``signed $Q1.6$" formats have similar trends and comparable accuracy as they represent similar weight values which differ only in the least significant fractional bit, thereby leading to similar neuron behavior and accuracy.   
These formats may have notable accuracy differences for some cases, such as after retraining process, as shown by~\circledB{6}. 
The possible reason is that these formats have different bit position for sign, integer, and fraction, thereby making the DRAM errors affect different weight bits and lead to different learning qualities during the respective fault-aware training.

In summary, our EnforceSNN maintains accuracy (i.e., no accuracy loss) as compared to the baseline with accurate DRAM when $ BER \leq 10^{\minus3}$ across different datasets.
Meanwhile, for higher BER values (i.e., $10^{\minus3} < BER \leq 10^{\minus2}$), our EnforceSNN still achieves higher accuracy than the baseline with accurate DRAM across different datasets.
Therefore, these results show that \textit{our EnforceSNN framework effectively improves the SNN error tolerance against DRAM errors with minimum retraining efforts}.

%%%%%%%%%%%%%%%%%%%%%%%
\subsection{DRAM Access Energy Savings and Throughput Improvements}
\label{Sec_Res_DRAMenergy}

\begin{figure}[t]
\centering
\includegraphics[width=\linewidth]{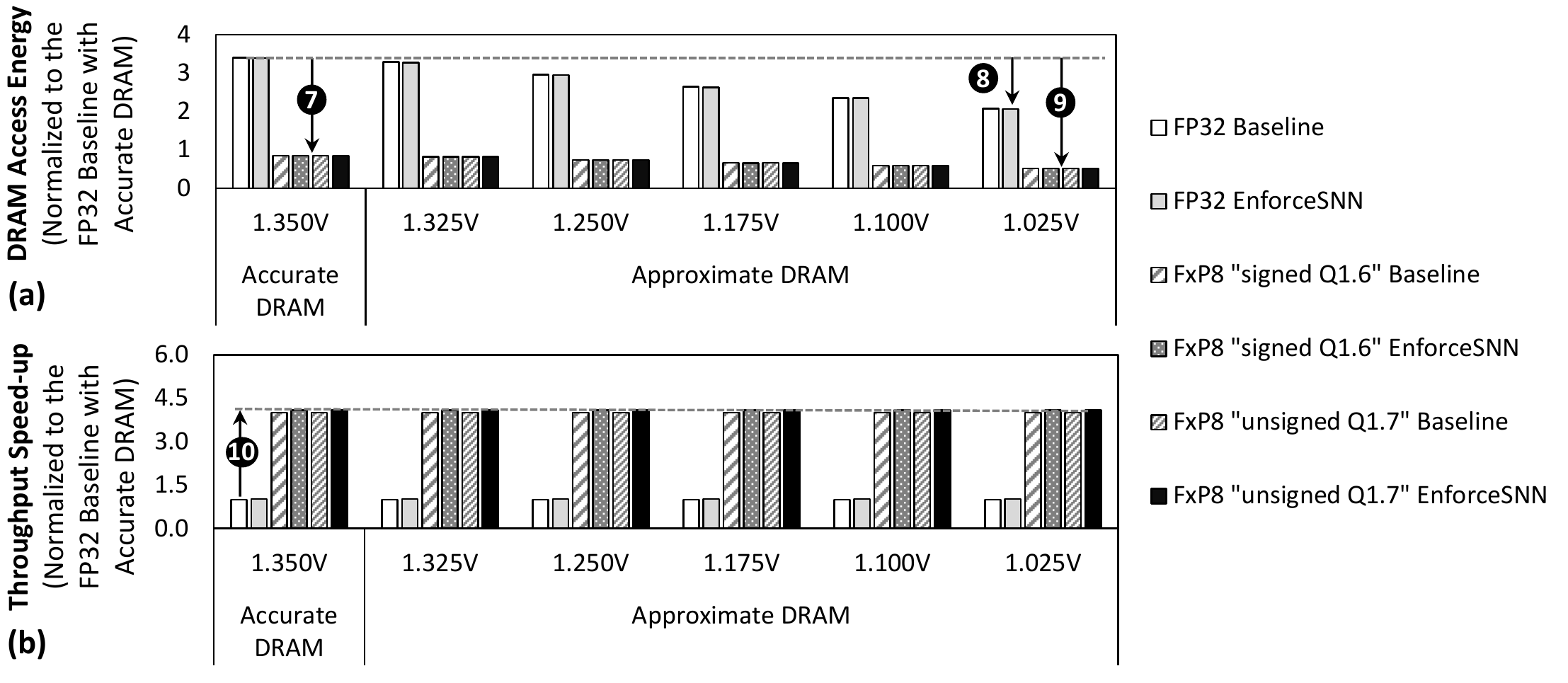}
\caption{(a) The normalized DRAM access energy for an inference incurred by the baseline model and the EnforceSNN-improved model with accurate and approximate DRAM, and (b) the normalized speed-up of DRAM data throughput for an inference achieved by our EnforceSNN-improved model over the baseline model, across different $V_{supply}$ values, different workloads (datasets), and different network sizes (N-900, N-1600, N-2500, and N-3600). These results are applicable for all network sizes. They are also applicable for both the MNIST and Fashion MNIST datasets, as these workloads have similar DRAM access energy, due to the same number of weights and number of DRAM accesses for an inference.}
\label{Fig_Results_EnergySpeed}
\end{figure}

Fig.~\ref{Fig_Results_EnergySpeed}(a) shows the normalized energy consumption of the DRAM accesses for an inference (i.e., inferring one input sample) required by the baseline model and the EnforceSNN-improved model with accurate and approximate DRAM, across different $V_{supply}$ values, precision levels, network sizes, and workloads.
We observe that different network sizes show similar normalized DRAM access energy, hence we only show a single figure representing the experimental results for all network sizes. 
For the accurate DRAM cases across different network sizes, the baseline model achieves 75\% DRAM energy saving when it employs the quantization technique, while our EnforceSNN-improved model achieves 75.1\% DRAM energy saving due to the quantization and the proposed DRAM mapping policy, as shown in~\circledB{7}.
Meanwhile, the difference in these DRAM energy savings comes from the DRAM mapping policy. 
That is, our EnforceSNN optimizes the DRAM energy-per-access by maximizing the row buffer hits and the multi-bank burst feature, thereby having fewer row buffer conflicts than the baseline which only exploits the single-bank burst feature. 
For the FP32 precision across different network sizes, employing the approximate DRAM in the baseline model reduces the DRAM energy savings by up to 39.2\% as compared to employing the accurate DRAM.
Meanwhile, employing the approximate DRAM in the EnforceSNN-improved model reduces the DRAM energy savings by up to 39.5\% as compared to employing the accurate DRAM, as shown in~\circledB{8}.
These energy savings come from the reduced DRAM energy-per-access due to the reduction of operational $V_{supply}$.
Moreover, the difference in energy savings between the baseline and our EnforceSNN also comes from the DRAM mapping policy.
For the FxP8 precision (i.e., ``signed $Q1.6$" and ``unsigned $Q1.7$") across different network sizes, employing the approximate DRAM in the baseline model reduces the DRAM energy savings by up to 84.8\% over employing the accurate one.
Meanwhile, employing the approximate DRAM in the EnforceSNN-improved model reduces the DRAM energy savings by up to 84.9\% over employing the accurate one, as shown in~\circledB{9}.
These energy savings come from the reduced weight precision and the reduced DRAM energy-per-access due to $V_{supply}$ reduction. 
Moreover, the difference in energy savings between the baseline and our EnforceSNN also comes from the DRAM mapping policy.
Furthermore, we also observe that our EnforceSNN-improved model obtains 4.1x throughput speed-up over the baseline model across different $V_{supply}$ values, workloads, and network sizes; see \circledB{10} in Fig.~\ref{Fig_Results_EnergySpeed}(b). 
It is achieved through (1) the quantization technique which reduces the number of DRAM accesses, and (2) our proposed DRAM mapping policy which optimizes the DRAM latency-per-access by maximizing the row buffer hits and the multi-bank burst features.
The results also show that the ``unsigned $Q1.7$" and ``signed $Q1.6$" achieve comparable DRAM access energy savings and throughput improvements since they employ the same bitwidth of weights, thereby having similar DRAM access behavior.

In summary, the results in Fig.~\ref{Fig_Results_EnergySpeed} indicate that \textit{our EnforceSNN framework substantially reduces the DRAM access energy by employing the reduced-voltage approximate DRAM and our efficient DRAM mapping policy, while effectively improving the DRAM data throughput mainly due to the quantization}.

%%%%%%%%%%%%%%%%%%%%%%%
\subsection{Model Selection under Design Trade-Offs}
\label{Sec_Res_TradeOffs}

\begin{figure}[t]
\centering
\includegraphics[width=0.81\linewidth]{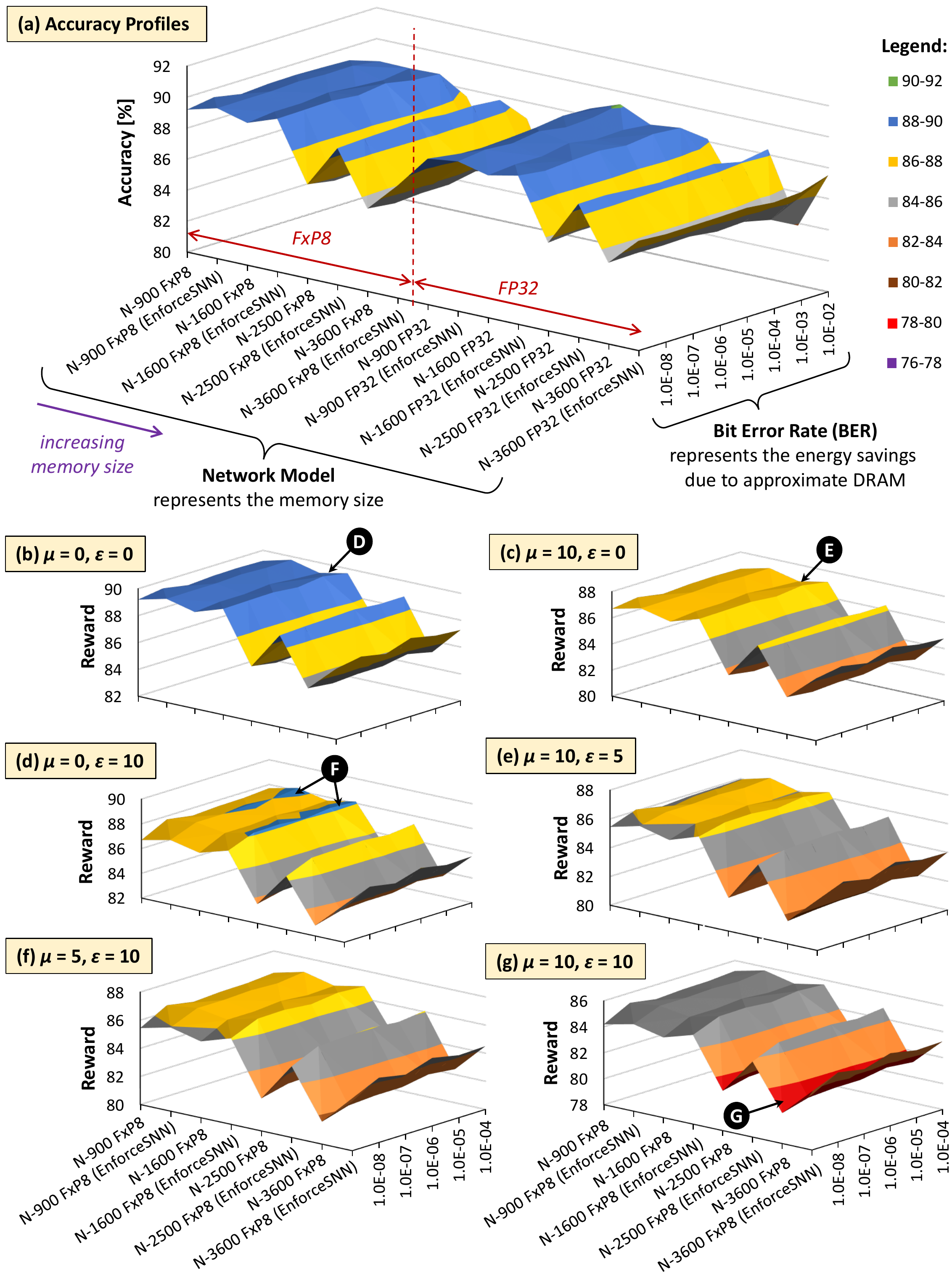}
\vspace{-0.3cm}
\caption{The trade-offs among accuracy, memory footprint, and energy consumption for the MNIST. (a) Accuracy profiles of SNN models. (b-g) Reward profiles of SNN models. The network sizes represent the memory sizes, and the BER values represent the energy savings from approximate DRAM.}
\label{Fig_Results_ModelSelect_MNIST}
\vspace{-0.3cm}
\end{figure}

\begin{figure}[t]
\centering
\includegraphics[width=0.81\linewidth]{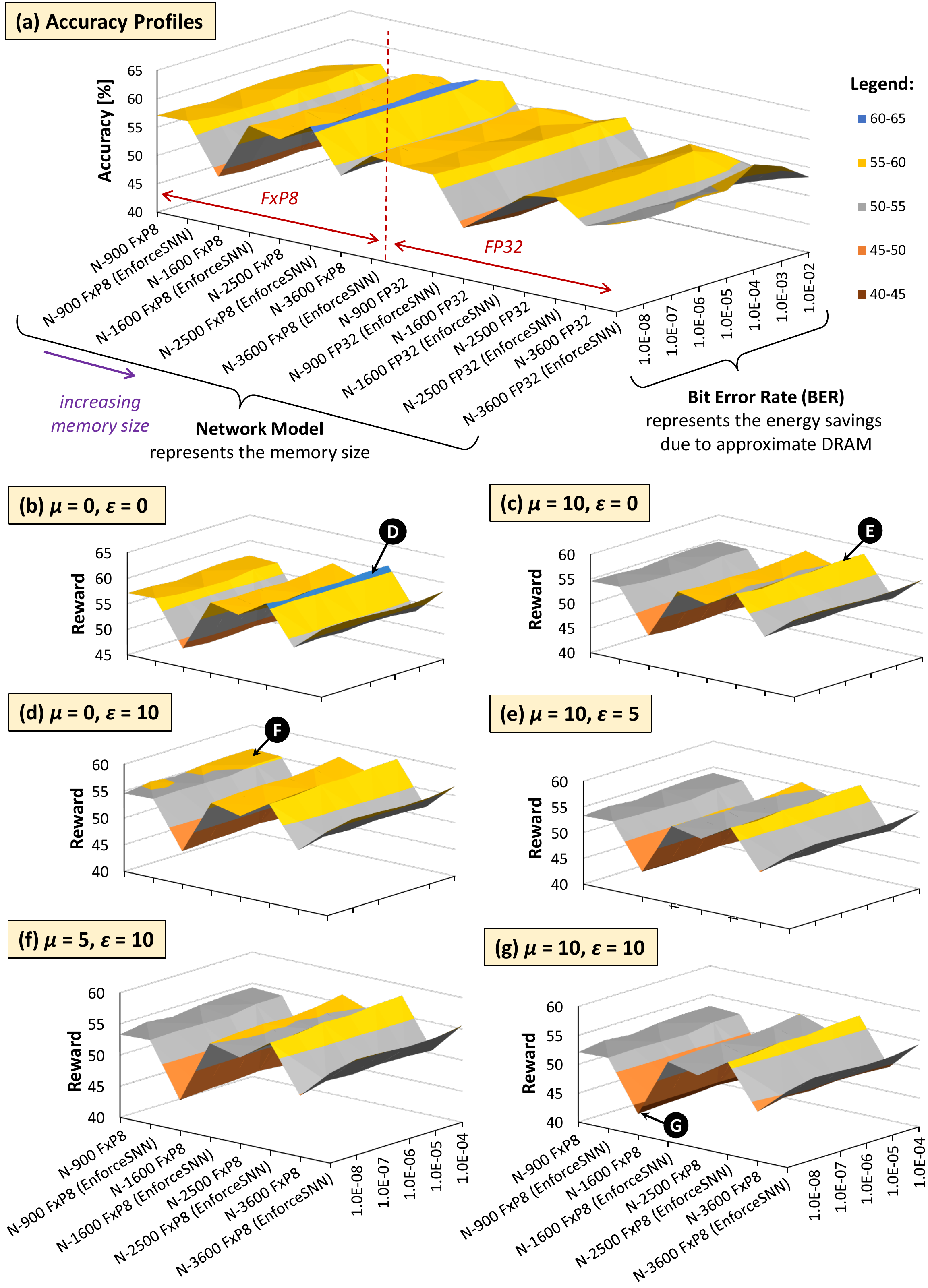}
\vspace{-0.3cm}
\caption{The trade-offs among accuracy, memory, and energy consumption for the Fashion MNIST. (a) Accuracy profiles of SNN models. (b-g) Reward profiles of SNN models. The network sizes represent the memory sizes, and the BER values represent the energy savings from approximate DRAM.}
\label{Fig_Results_ModelSelect_FMNIST}
\vspace{-0.3cm}
\end{figure}

Fig.~\ref{Fig_Results_ModelSelect_MNIST} and Fig.~\ref{Fig_Results_ModelSelect_FMNIST} show the results of the accuracy-memory-energy trade-offs for the MNIST and Fashion MNIST datasets, respectively.
In this evaluation, the quantized models consider the FxP8 precision with ``signed $Q1.6$" format.
For the given SNN model candidates, we observe that the models that incur small memory size typically employ FxP8 precision, as shown in Fig.~\ref{Fig_Results_ModelSelect_MNIST}(a) for the MNIST and Fig.~\ref{Fig_Results_ModelSelect_FMNIST}(a) for the Fashion MNIST.
Considering that the accuracy of the FxP8-based models is comparable to the FP32-based models, we narrow down the candidates to only the FxP8-based models.

To analyze the design trade-offs, we explore the impact of different $\mu$ and $\varepsilon$ values on the rewards. 
For instance, if we consider that the accuracy should have a higher priority than the memory and energy consumption, we set $\mu$ and $\varepsilon$ low (e.g., $\mu = 0$ and $\varepsilon = 0$).
Meanwhile, if we consider that the memory should have a higher priority than the energy consumption, we set $\mu$ higher than $\varepsilon$ (e.g., $\mu = 10$ and $\varepsilon = 0$).
For both cases, the highest reward is achieved by the EnforceSNN-improved N-1600 FxP8 for the MNIST and the EnforceSNN-improved N-2500 FxP8 for the Fashion MNIST under $10^{-5}$ error rate; see~\circledB{D} for $\mu = 0$ and $\varepsilon = 0$ case, and see~\circledB{E} for $\mu = 10$ and $\varepsilon = 0$ case.
The reason is that, these models employ our efficient FAT technique to improve their error tolerance, thereby leading to high accuracy under a high error rate. 
We also observe that having $\mu$ higher than $\varepsilon$ makes the high rewards shift towards smaller models, as shown by~\circledB{E} in Fig.~\ref{Fig_Results_ModelSelect_MNIST}(c). 
The reason is that a higher $\mu$ makes the small $m_{norm}$ have a smaller impact on the reward reduction than the large $m_{norm}$, thereby maintaining the high reward values. 
If the energy consumption should have a higher priority than the memory footprint, we set $\mu$ lower than $\varepsilon$ (e.g., $\mu = 0$ and $\varepsilon = 10$). 
The highest reward is achieved by the EnforceSNN-improved N-1600 FxP8 under $10^{-5}$ error rate for the MNIST and the EnforceSNN-improved N-2500 FxP8 under $10^{-4}$ error rate for the Fashion MNIST. 
In this case, we observe that high rewards are shifted towards models with smaller energy consumption (represented by higher BER); see~\circledB{F} in Fig.~\ref{Fig_Results_ModelSelect_MNIST}(d) and Fig.~\ref{Fig_Results_ModelSelect_FMNIST}(d).
The reason is that a higher $\varepsilon$ makes the small $E_{norm}$ have a smaller impact on the reward reduction than the large $E_{norm}$, thereby maintaining the high reward values. 
Furthermore, if the memory and energy consumption should have a higher priority than the accuracy, we set $\mu$ and $\varepsilon$ high (e.g., $\mu = 10$ and $\varepsilon = 10$). 
The highest reward is achieved by the EnforceSNN-improved N-1600 FxP8 under $10^{-5}$ error rate for the MNIST and the EnforceSNN-improved N-2500 FxP8 under $10^{-4}$ error rate for the Fashion MNIST. 
In this case, high rewards are shifted towards models with smaller memory and energy consumption (represented by high BER), but their overall rewards decrease as the values of $\mu$ and $\varepsilon$ increase; see~\circledB{G} in Fig.~\ref{Fig_Results_ModelSelect_MNIST}(e) and Fig.~\ref{Fig_Results_ModelSelect_FMNIST}(e).
The reason is that, higher $\mu$ and $\varepsilon$ jointly make the $m_{norm}$ and $E_{norm}$ decrease the reward. 
It means that if we want to significantly reduce the memory footprint and energy consumption, we have to accept more accuracy degradation. 

In summary, the results in Fig.~\ref{Fig_Results_ModelSelect_MNIST} and Fig.~\ref{Fig_Results_ModelSelect_FMNIST} show that \textit{our EnforceSNN framework has an effective algorithm to trade off the accuracy, memory footprint, and energy consumption of the given SNN models}, thereby providing good applicability for diverse embedded applications with their respective constraints.

%%%%%%%%%%%%%%%%%%%%%%%
\subsection{Optimization of the Retraining Costs}
\label{Sec_Res_Retrain}

The conventional FAT for neural networks usually injects errors at an incremental rate during the retraining process from the minimum value to the maximum one for avoiding accuracy collapse~\citep{Ref_Koppula_EDEN_MICRO19}. 
Therefore, in this work, the conventional FAT considers $BER=\{10^{-8}, 10^{-7}, 10^{-6}, ..., 10^{-2}\}$, while our efficient FAT in EnforceSNN only considers $BER=\{10^{-4}, 10^{-3}, 10^{-2}\}$ in the retraining process.

\begin{figure}[t]
\centering
\includegraphics[width=\linewidth]{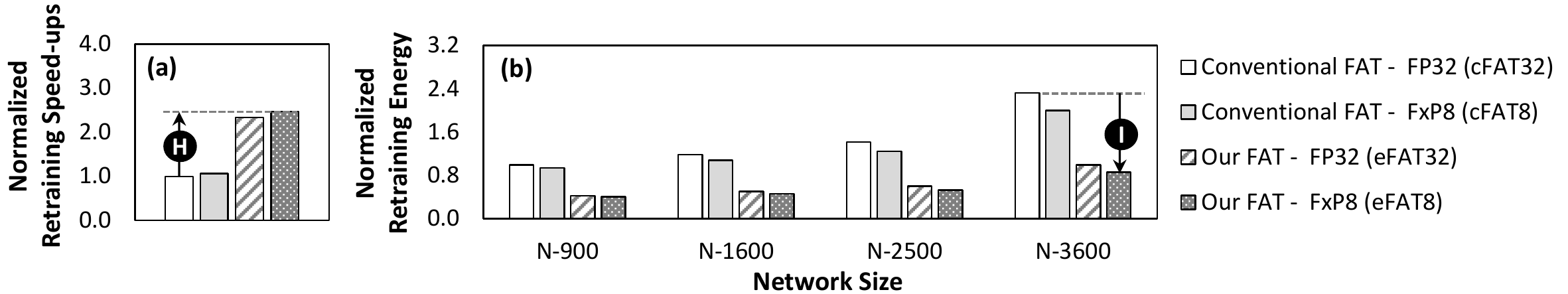}
\vspace{-0.3cm}
\caption{(a) The retraining speed-ups across different network sizes (i.e., N-900, N-1600, N-2500, and N-3600), and (b) the retraining energy for the MNIST, which are normalized to the conventional FAT with FP32 for a 900-neuron network. 
The results for the Fashion MNIST show similar trends to the MNIST since these workloads have similar DRAM access latency and energy due to the same number of weights and number of DRAM accesses for a training phase.
Here, FxP8 represents both the ``signed $Q1.6$" and ``unsigned $Q1.7$" formats.}
\label{Fig_Results_Retrain}
\end{figure}

\textbf{Retraining Speed-ups:} 
The conventional FAT with FxP8 (cFAT8) obtains speed-up over the one with FP32 (cFAT32) by up to 1.16x and 1.14x for the MNIST and the Fashion MNIST respectively, since the cFAT8 employs quantized weights, thereby leading to a faster error injection and learning process.   
Meanwhile, our efficient FAT with FP32 (eFAT32) obtains a 2.33x speed-up over the cFAT32, since our eFAT32 has fewer iterations of retraining process.
Furthermore, we also observe that our efficient FAT with FxP8 (eFAT8) obtains more speed-up, i.e., by up to 2.71x for the MNIST and 2.65x for the Fashion MNIST as shown by~\circledB{H} in Fig.~\ref{Fig_Results_Retrain}(a), since our eFAT8 employs quantized weights in addition to fewer iterations of the retraining process.

\textbf{Retraining Energy Savings:} 
The cFAT8 achieves energy saving over the cFAT32 by up to 13.9\% for the MNIST and 12\% for the Fashion MNIST, since the cFAT8 employs quantized weights which incur lower energy consumption during the error injection and learning process.   
Meanwhile, our eFAT32 achieves energy saving over the cFAT32 by 57.1\%, as the eFAT32 performs fewer iterations of the retraining process as compared to the cFAT32.
Our eFAT8 achieves further energy saving, i.e., by up to 63.1\% for the MNIST and by up to 62.3\% for the Fashion MNIST as shown by~\circledB{I} in Fig.~\ref{Fig_Results_Retrain}(b), since it employs quantized weights in addition to fewer iterations of the retraining process, thereby leading to a higher energy saving.

\textbf{Carbon Emission Reduction:}
The retraining process also poses additional challenges that correspond to environmental concerns, i.e., carbon emission.
Recent works have highlighted that the carbon emission from neural network training should be minimized to prevent the increasing rates of natural disasters~\citep{Ref_Strubell_Carbon_ACL19,Ref_Strubell_Carbon_AAAI20}.
To estimate the carbon emission of neural network training, the work of~\citet{Ref_Strubell_Carbon_ACL19} proposed Eq.~\ref{Eq_CO2e} and Eq.~\ref{Eq_pt}.
In these equations, $CO_2e$ denotes the estimated carbon ($CO_2$) emission during the training, which is a function of the total power during the training ($p_t$). 
Meanwhile, $t$ is the training duration, $p_c$ is the average power from all CPUs, $p_r$ is the average power from all main memories (DRAMs), $p_g$ is the average power from a GPU and $g$ is the number of GPUs.
\begin{equation}
CO_2e = 0.954 \cdot p_t
\label{Eq_CO2e}
\end{equation}
\begin{equation}
p_t = \frac{1.58 \cdot t (p_c+p_r + g \cdot p_g)}{1000}
\label{Eq_pt}
\end{equation}

These equations indicate that if we assume $p_c$, $p_r$, $p_g$, and $g$ are the same for different FAT techniques, then the difference will come from the training duration $t$.  
Therefore, our efficient FAT in EnforceSNN employs fewer iterations of the retraining process than the conventional FAT, thereby producing less carbon emission. 
Moreover, our EnforceSNN also reduces the operational power of the main memory (DRAM) through the reduced-voltage approximation approach, thereby further reducing the emission.

In summary, \textit{our EnforceSNN framework effectively offers speed-up of retraining time, reduction of retraining energy, and less carbon emission than the conventional FAT technique}, thereby making it more friendly for our environments.

%%%%%%%%%%%%%%%%%%%%%%%
\subsection{Further Discussion}
\label{Sec_Res_Further}

Previous works that exploit the reduced-voltage DRAM concept mainly aim at improving the energy efficiency of mobile systems~\citep{Ref_HajYahya_SysScale_ISCA20}, personal computing systems~\citep{Ref_Larimi_HBMvolt_DATE21, Ref_FabricioFilho_SmartApprox_SusCom22}, and server systems~\citep{Ref_Larimi_HBMvolt_DATE21, Ref_David_MemDVFS_ICAC11, Ref_Deng_MemScale_ASPLOS11, Ref_Deng_CoScale_Micro12, Ref_Deng_MultiScale_ISLPED12}. 
This concept is also employed for minimizing the energy consumption of deep neural networks (DNNs)~\citep{Ref_Koppula_EDEN_MICRO19}. 
Since SNNs have different data representation, computation models, and learning rules as compared to DNNs, our EnforceSNN provides a different framework with different techniques that are crafted specifically for improving the resilience and the energy efficiency of SNNs.  
Furthermore, the reduced-voltage DRAM is also used to generate noise (i.e., from DRAM errors) for obfuscating the intellectual property (IP) against security threats, such as IP stealing~\citep{Ref_Xu_MIDAS_AsianHOST20}.

Our EnforceSNN framework can be put in the approximate computing field, especially in the context of the approximation for main memory through voltage scaling~\citep{Ref_Venkataramani_ApproxCom_DAC15, Ref_Xu_ApproxCompSurvey_MDAT16, Ref_Mittal_ApproxComSurvey_CSUR16}.
Therefore, some of the techniques in our EnforceSNN are suitable for different domains outside SNNs: (1) DRAM voltage reduction for optimizing the DRAM access energy, (2) quantization for reducing the memory footprint, and (3) error-aware DRAM data mapping policy for minimizing the negative impact of DRAM errors to the data. 
These techniques are applicable for error-tolerant applications, such as image/video processing (e.g., data compression) and data analytic applications (e.g., data clustering).

%%%%%%%%%%%%%%%%%%%%%%%%%%%%%%%%%%%%%%%%%%%%%%%%%%%%%%%
%%%%%%%%%%%%%%%%%%%%%%%%%%%%%%%%%%%%%%%%%%%%%%%%%%%%%%%
\section{Conclusion}
\label{Sec_Conclude}

We propose a novel EnforceSNN framework to achieve resilient
and energy-efficient SNN inference considering reduced-voltage-based approximate DRAM, through weight quantization, error-aware DRAM mapping, SNN error-tolerance analysis, efficient error-aware SNN training, and effective SNN model selection. 
Our EnforceSNN achieves no accuracy loss for BER $\leq 10^{-3}$ with minimum retraining costs as compared to the baseline SNN with accurate DRAM while achieving up to 84.9\% of DRAM energy saving and up to 4.1x speed-up of DRAM data throughput. 
In this manner, our work may enable efficient SNN inference for energy-constrained embedded devices like the Edge-AI.

%%%%%%%%%%%%%%%%%%%%%%%%%%%%%%%%%%%%%%%%%%%%%%%%%%%%%%%
%%%%%%%%%%%%%%%%%%%%%%%%%%%%%%%%%%%%%%%%%%%%%%%%%%%%%%%
\section*{Data Availability Statement}
\label{Sec_DataStatement}

Publicly available datasets were used in this study. 
These datasets can be found at the following sites: http://yann.lecun.com/exdb/mnist/ (MNIST),
http://fashion-mnist.s3-website.eu-central-1.amazonaws.com/ (Fashion MNIST).

%%%%%%%%%%%%%%%%%%%%%%%%%%%%%%%%%%%%%%%%%%%%%%%%%%%%%%%
%%%%%%%%%%%%%%%%%%%%%%%%%%%%%%%%%%%%%%%%%%%%%%%%%%%%%%%
\section*{Funding}
\label{Sec_Funding}

This research is partly supported by the ASPIRE AARE Grant (S1561) on ``Towards Extreme Energy Efficiency through Cross-Layer Approximate Computing".

%%%%%%%%%%%%%%%%%%%%%%%%%%%%%%%%%%%%%%%%%%%%%%%%%%%%%%%
%%%%%%%%%%%%%%%%%%%%%%%%%%%%%%%%%%%%%%%%%%%%%%%%%%%%%%%

% \bibliographystyle{Frontiers-Harvard}
\bibliographystyle{frontiersinSCNS_ENG_HUMS} % for Science, Engineering and Humanities and Social Sciences articles, for Humanities and Social Sciences articles please include page numbers in the in-text citations
\bibliography{Bibliography}

\begin{thebibliography}{70}
\providecommand{\natexlab}[1]{#1}
\expandafter\ifx\csname urlstyle\endcsname\relax
  \providecommand{\doi}[1]{doi:\discretionary{}{}{}#1}\else
  \providecommand{\doi}{doi:\discretionary{}{}{}\begingroup
  \urlstyle{rm}\Url}\fi
\providecommand{\selectlanguage}[1]{\relax}
\providecommand{\bibAnnoteFile}[1]{%
  \IfFileExists{#1}{\begin{quotation}\noindent\textsc{Key:} #1\\
  \textsc{Annotation:}\ \input{#1}\end{quotation}}{}}
\providecommand{\bibAnnote}[2]{%
  \begin{quotation}\noindent\textsc{Key:} #1\\
  \textsc{Annotation:}\ #2\end{quotation}}

\bibitem[{{Akopyan} et~al.(2015){Akopyan}, {Sawada}, {Cassidy},
  {Alvarez-Icaza}, {Arthur}, {Merolla} et~al.}]{Ref_Akopyan_TrueNorth_TCAD15}
{Akopyan}, F., {Sawada}, J., {Cassidy}, A., {Alvarez-Icaza}, R., {Arthur}, J.,
  {Merolla}, P., et~al. (2015).
\newblock \uppercase{T}rue\uppercase{N}orth: Design and tool flow of a 65 mw 1
  million neuron programmable neurosynaptic chip.
\newblock \emph{IEEE Transactions on Computer-Aided Design of Integrated
  Circuits and Systems} 34.
\newblock \doi{10.1109/TCAD.2015.2474396}
\bibAnnoteFile{Ref_Akopyan_TrueNorth_TCAD15}

\bibitem[{Cao et~al.(2020)Cao, Liu, Meng, and Sun}]{Ref_Cao_Edge_Access20}
Cao, K., Liu, Y., Meng, G., and Sun, Q. (2020).
\newblock An overview on edge computing research.
\newblock \emph{IEEE Access} 8, 85714--85728.
\newblock \doi{10.1109/ACCESS.2020.2991734}
\bibAnnoteFile{Ref_Cao_Edge_Access20}

\bibitem[{Chandrasekar(2014)}]{Ref_Chandrasekar_DRAMPower}
Chandrasekar, K. (2014).
\newblock \emph{High-level power estimation and optimization of DRAMs}.
\newblock Ph.D. thesis, TU Delft
\bibAnnoteFile{Ref_Chandrasekar_DRAMPower}

\bibitem[{Chang et~al.(2017)Chang, Ya\u{g}l\i{}k\c{c}\i{}, Ghose, Agrawal,
  Chatterjee, Kashyap et~al.}]{Ref_Chang_Voltron_POMACS17}
Chang, K.~K., Ya\u{g}l\i{}k\c{c}\i{}, A.~G., Ghose, S., Agrawal, A.,
  Chatterjee, N., Kashyap, A., et~al. (2017).
\newblock \uppercase{U}nderstanding reduced-voltage operation in modern
  \uppercase{DRAM} devices: Experimental characterization, analysis, and
  mechanisms.
\newblock \emph{Proc. ACM on Measurements and Analysis of Computing Systems} 1.
\newblock \doi{10.1145/3084447}
\bibAnnoteFile{Ref_Chang_Voltron_POMACS17}

\bibitem[{Chattopadhyay et~al.(2017)Chattopadhyay, Prakash, and
  Shafique}]{Ref_Chattopadhyay_CPS_DATE17}
Chattopadhyay, A., Prakash, A., and Shafique, M. (2017).
\newblock Secure cyber-physical systems: Current trends, tools and open
  research problems.
\newblock In \emph{Design, Automation \& Test in Europe Conference \&
  Exhibition, 2017}. 1104--1109.
\newblock \doi{10.23919/DATE.2017.7927154}
\bibAnnoteFile{Ref_Chattopadhyay_CPS_DATE17}

\bibitem[{Chen and Ran(2019)}]{Ref_Chen_DLwithEdge_JPROC19}
Chen, J. and Ran, X. (2019).
\newblock Deep learning with edge computing: A review.
\newblock \emph{Proceedings of the IEEE} 107, 1655--1674.
\newblock \doi{10.1109/JPROC.2019.2921977}
\bibAnnoteFile{Ref_Chen_DLwithEdge_JPROC19}

\bibitem[{David et~al.(2011)David, Fallin, Gorbatov, Hanebutte, and
  Mutlu}]{Ref_David_MemDVFS_ICAC11}
David, H., Fallin, C., Gorbatov, E., Hanebutte, U.~R., and Mutlu, O. (2011).
\newblock Memory power management via dynamic voltage/frequency scaling.
\newblock In \emph{The 8th ACM International Conference on Autonomic
  Computing}. 31–40.
\newblock \doi{10.1145/1998582.1998590}
\bibAnnoteFile{Ref_David_MemDVFS_ICAC11}

\bibitem[{Deng et~al.(2012{\natexlab{a}})Deng, Meisner, Bhattacharjee, Wenisch,
  and Bianchini}]{Ref_Deng_CoScale_Micro12}
Deng, Q., Meisner, D., Bhattacharjee, A., Wenisch, T.~F., and Bianchini, R.
  (2012{\natexlab{a}}).
\newblock Co\uppercase{S}cale: Coordinating \uppercase{CPU} and memory system
  \uppercase{DVFS} in server systems.
\newblock In \emph{2012 45th Annual IEEE/ACM International Symposium on
  Microarchitecture}. 143--154.
\newblock \doi{10.1109/MICRO.2012.22}
\bibAnnoteFile{Ref_Deng_CoScale_Micro12}

\bibitem[{Deng et~al.(2012{\natexlab{b}})Deng, Meisner, Bhattacharjee, Wenisch,
  and Bianchini}]{Ref_Deng_MultiScale_ISLPED12}
Deng, Q., Meisner, D., Bhattacharjee, A., Wenisch, T.~F., and Bianchini, R.
  (2012{\natexlab{b}}).
\newblock Multi\uppercase{S}cale: Memory system \uppercase{DVFS} with multiple
  memory controllers.
\newblock In \emph{The 2012 ACM/IEEE International Symposium on Low Power
  Electronics and Design}. 297–302.
\newblock \doi{10.1145/2333660.2333727}
\bibAnnoteFile{Ref_Deng_MultiScale_ISLPED12}

\bibitem[{Deng et~al.(2011)Deng, Meisner, Ramos, Wenisch, and
  Bianchini}]{Ref_Deng_MemScale_ASPLOS11}
Deng, Q., Meisner, D., Ramos, L., Wenisch, T.~F., and Bianchini, R. (2011).
\newblock Mem\uppercase{S}cale: Active low-power modes for main memory.
\newblock In \emph{The 16th International Conference on Architectural Support
  for Programming Languages and Operating Systems}. 225–238.
\newblock \doi{10.1145/1950365.1950392}
\bibAnnoteFile{Ref_Deng_MemScale_ASPLOS11}

\bibitem[{Diehl and Cook(2015)}]{Ref_Diehl_STDPmnist_FNCOM15}
Diehl, P. and Cook, M. (2015).
\newblock Unsupervised learning of digit recognition using
  spike-timing-dependent plasticity.
\newblock \emph{Frontiers in Computational Neuroscience} 9, 99.
\newblock \doi{10.3389/fncom.2015.00099}
\bibAnnoteFile{Ref_Diehl_STDPmnist_FNCOM15}

\bibitem[{{Fabrício Filho} et~al.(2022){Fabrício Filho}, Felzmann, and
  Wanner}]{Ref_FabricioFilho_SmartApprox_SusCom22}
{Fabrício Filho}, J., Felzmann, I., and Wanner, L. (2022).
\newblock Smart\uppercase{A}pprox: Learning-based configuration of approximate
  memories for energy-efficient execution.
\newblock \emph{Sustainable Computing: Informatics and Systems} 34, 100701.
\newblock \doi{10.1016/j.suscom.2022.100701}
\bibAnnoteFile{Ref_FabricioFilho_SmartApprox_SusCom22}

\bibitem[{{Frenkel} et~al.(2019){Frenkel}, {Lefebvre}, {Legat}, and
  {Bol}}]{Ref_Frenkel_ODIN_TBCAS19}
{Frenkel}, C., {Lefebvre}, M., {Legat}, J., and {Bol}, D. (2019).
\newblock A 0.086-mm$^2$ 12.7-pj/sop 64k-synapse 256-neuron online-learning
  digital spiking neuromorphic processor in 28-nm cmos.
\newblock \emph{IEEE Transactions on Biomedical Circuits and Systems} 13,
  145--158.
\newblock \doi{10.1109/TBCAS.2018.2880425}
\bibAnnoteFile{Ref_Frenkel_ODIN_TBCAS19}

\bibitem[{Frenkel et~al.(2019)Frenkel, Legat, and
  Bol}]{Ref_Frenkel_MorphIC_TBCAS19}
Frenkel, C., Legat, J.-D., and Bol, D. (2019).
\newblock Morph\uppercase{IC}: A 65-nm 738k-synapse/mm$^2$ quad-core
  binary-weight digital neuromorphic processor with stochastic spike-driven
  online learning.
\newblock \emph{IEEE Transactions on Biomedical Circuits and Systems} 13,
  999--1010.
\newblock \doi{10.1109/TBCAS.2019.2928793}
\bibAnnoteFile{Ref_Frenkel_MorphIC_TBCAS19}

\bibitem[{Gautrais and Thorpe(1998)}]{Ref_Gautrais_SpikeCoding_Bio98}
Gautrais, J. and Thorpe, S. (1998).
\newblock Rate coding versus temporal order coding: a theoretical approach.
\newblock \emph{Biosystems} 48, 57--65.
\newblock \doi{10.1016/S0303-2647(98)00050-1}
\bibAnnoteFile{Ref_Gautrais_SpikeCoding_Bio98}

\bibitem[{Gholami et~al.(2021)Gholami, Kim, Dong, Yao, Mahoney, and
  Keutzer}]{Ref_Gholami_QuantSurvey_arXiv21}
Gholami, A., Kim, S., Dong, Z., Yao, Z., Mahoney, M.~W., and Keutzer, K.
  (2021).
\newblock A survey of quantization methods for efficient neural network
  inference.
\newblock \emph{arXiv preprint arXiv:2103.13630}
\bibAnnoteFile{Ref_Gholami_QuantSurvey_arXiv21}

\bibitem[{Ghose et~al.(2019)Ghose, Li, Hajinazar, Cali, and
  Mutlu}]{Ref_Ghose_WorkloadDRAM_POMACS19}
Ghose, S., Li, T., Hajinazar, N., Cali, D.~S., and Mutlu, O. (2019).
\newblock Demystifying complex workload-\uppercase{DRAM} interactions: An
  experimental study.
\newblock \emph{Proc. ACM on Measurements and Analysis of Computing Systems} 3.
\newblock \doi{10.1145/3366708}
\bibAnnoteFile{Ref_Ghose_WorkloadDRAM_POMACS19}

\bibitem[{Griffor et~al.(2017)Griffor, Greer, Wollman, and
  Burns}]{Ref_Griffor_CPS_NIST17}
Griffor, E., Greer, C., Wollman, D., and Burns, M. (2017).
\newblock \emph{Framework for \uppercase{C}yber-\uppercase{P}hysical
  \uppercase{S}ystems: Volume 1, \uppercase{O}verview} (Special Publication
  (NIST SP), National Institute of Standards and Technology, Gaithersburg, MD).
\newblock \doi{10.6028/NIST.SP.1500-201}
\bibAnnoteFile{Ref_Griffor_CPS_NIST17}

\bibitem[{Gupta et~al.(2015)Gupta, Agrawal, Gopalakrishnan, and
  Narayanan}]{Ref_Gupta_DLPrecision_ICML15}
Gupta, S., Agrawal, A., Gopalakrishnan, K., and Narayanan, P. (2015).
\newblock Deep learning with limited numerical precision.
\newblock In \emph{32nd International Conference on Machine Learning}.
  1737–1746
\bibAnnoteFile{Ref_Gupta_DLPrecision_ICML15}

\bibitem[{Haj-Yahya et~al.(2020)Haj-Yahya, Alser, Kim, Yağlıkçı,
  Vijaykumar, Rotem et~al.}]{Ref_HajYahya_SysScale_ISCA20}
Haj-Yahya, J., Alser, M., Kim, J., Yağlıkçı, A.~G., Vijaykumar, N., Rotem,
  E., et~al. (2020).
\newblock Sys\uppercase{S}cale: Exploiting multi-domain dynamic voltage and
  frequency scaling for energy efficient mobile processors.
\newblock In \emph{2020 ACM/IEEE 47th Annual International Symposium on
  Computer Architecture}. 227--240.
\newblock \doi{10.1109/ISCA45697.2020.00029}
\bibAnnoteFile{Ref_HajYahya_SysScale_ISCA20}

\bibitem[{{Hazan} et~al.(2018){Hazan}, {Saunders}, {Sanghavi}, {Siegelmann},
  and {Kozma}}]{Ref_Hazan_SOMSNN_IJCNN18}
{Hazan}, H., {Saunders}, D., {Sanghavi}, D.~T., {Siegelmann}, H., and {Kozma},
  R. (2018).
\newblock Unsupervised learning with self-organizing spiking neural networks.
\newblock In \emph{2018 International Joint Conference on Neural Networks}.
  1--6.
\newblock \doi{10.1109/IJCNN.2018.8489673}
\bibAnnoteFile{Ref_Hazan_SOMSNN_IJCNN18}

\bibitem[{Hazan et~al.(2018)Hazan, Saunders, Khan, Patel, Sanghavi, Siegelmann
  et~al.}]{Ref_Hazan_BindsNET_FNINF18}
Hazan, H., Saunders, D.~J., Khan, H., Patel, D., Sanghavi, D.~T., Siegelmann,
  H.~T., et~al. (2018).
\newblock Bindsnet: A machine learning-oriented spiking neural networks library
  in python.
\newblock \emph{Frontiers in Neuroinformatics} \doi{10.3389/fninf.2018.00089}
\bibAnnoteFile{Ref_Hazan_BindsNET_FNINF18}

\bibitem[{Hopkins et~al.(2020)Hopkins, Mikaitis, Lester, and
  Furber}]{Ref_Hopkins_Rounding_RSTA20}
Hopkins, M., Mikaitis, M., Lester, D.~R., and Furber, S. (2020).
\newblock Stochastic rounding and reduced-precision fixed-point arithmetic for
  solving neural ordinary differential equations.
\newblock \emph{Philosophical Transactions of the Royal Society A} 378,
  20190052.
\newblock \doi{10.1098/rsta.2019.0052}
\bibAnnoteFile{Ref_Hopkins_Rounding_RSTA20}

\bibitem[{{Izhikevich}(2004)}]{Ref_Izhikevich_CompareModels_TNN04}
{Izhikevich}, E.~M. (2004).
\newblock Which model to use for cortical spiking neurons?
\newblock \emph{IEEE Transactions on Neural Networks} 15, 1063--1070
\bibAnnoteFile{Ref_Izhikevich_CompareModels_TNN04}

\bibitem[{Jacob et~al.(2018)Jacob, Kligys, Chen, Zhu, Tang, Howard
  et~al.}]{Ref_Jacob_QuantNN_CVPR18}
Jacob, B., Kligys, S., Chen, B., Zhu, M., Tang, M., Howard, A., et~al. (2018).
\newblock Quantization and training of neural networks for efficient
  integer-arithmetic-only inference.
\newblock In \emph{The IEEE Conference on Computer Vision and Pattern
  Recognition}. 2704--2713
\bibAnnoteFile{Ref_Jacob_QuantNN_CVPR18}

\bibitem[{Kayser et~al.(2009)Kayser, Montemurro, Logothetis, and
  Panzeri}]{Ref_Kayser_PhaseCoding_Neuron09}
Kayser, C., Montemurro, M.~A., Logothetis, N.~K., and Panzeri, S. (2009).
\newblock Spike-phase coding boosts and stabilizes information carried by
  spatial and temporal spike patterns.
\newblock \emph{Neuron} 61, 597 -- 608.
\newblock \doi{10.1016/j.neuron.2009.01.008}
\bibAnnoteFile{Ref_Kayser_PhaseCoding_Neuron09}

\bibitem[{Kim et~al.(2018)Kim, Patel, Hassan, and
  Mutlu}]{Ref_Kim_SolarDRAM_ICCD18}
Kim, J., Patel, M., Hassan, H., and Mutlu, O. (2018).
\newblock \uppercase{S}olar-\uppercase{DRAM}: Reducing \uppercase{DRAM} access
  latency by exploiting the variation in local bitlines.
\newblock In \emph{2018 IEEE 36th International Conference on Computer Design}.
  282--291.
\newblock \doi{10.1109/ICCD.2018.00051}
\bibAnnoteFile{Ref_Kim_SolarDRAM_ICCD18}

\bibitem[{Kim et~al.(2012)Kim, Seshadri, Lee, Liu, and
  Mutlu}]{Ref_Kim_SALP_ISCA12}
Kim, Y., Seshadri, V., Lee, D., Liu, J., and Mutlu, O. (2012).
\newblock \uppercase{A} case for exploiting subarray-level parallelism
  (\uppercase{SALP}) in \uppercase{DRAM}.
\newblock In \emph{2012 39th Annual International Symposium on Computer
  Architecture}. 368--379.
\newblock \doi{10.1109/ISCA.2012.6237032}
\bibAnnoteFile{Ref_Kim_SALP_ISCA12}

\bibitem[{Koppula et~al.(2019)Koppula, Orosa, Ya\u{g}l\i{}k\c{c}\i{}, Azizi,
  Shahroodi, Kanellopoulos et~al.}]{Ref_Koppula_EDEN_MICRO19}
Koppula, S., Orosa, L., Ya\u{g}l\i{}k\c{c}\i{}, A.~G., Azizi, R., Shahroodi,
  T., Kanellopoulos, K., et~al. (2019).
\newblock \uppercase{EDEN}: Enabling energy-efficient, high-performance deep
  neural network inference using approximate \uppercase{DRAM}.
\newblock In \emph{52nd Annual IEEE/ACM International Symposium on
  Microarchitecture}. 166–181.
\newblock \doi{10.1145/3352460.3358280}
\bibAnnoteFile{Ref_Koppula_EDEN_MICRO19}

\bibitem[{Kriebel et~al.(2018)Kriebel, Rehman, Hanif, Khalid, and
  Shafique}]{Ref_Kriebel_CPSIoT_ISVLSI18}
Kriebel, F., Rehman, S., Hanif, M.~A., Khalid, F., and Shafique, M. (2018).
\newblock Robustness for smart cyber physical systems and internet-of-things:
  From adaptive robustness methods to reliability and security for machine
  learning.
\newblock In \emph{2018 IEEE Computer Society Annual Symposium on VLSI}.
  581--586.
\newblock \doi{10.1109/ISVLSI.2018.00111}
\bibAnnoteFile{Ref_Kriebel_CPSIoT_ISVLSI18}

\bibitem[{Krishnamoorthi(2018)}]{Ref_Krishnamoorthi_Whitepaper_arXiv18}
Krishnamoorthi, R. (2018).
\newblock Quantizing deep convolutional networks for efficient inference: A
  whitepaper.
\newblock \emph{arXiv preprint arXiv:1806.08342}
\bibAnnoteFile{Ref_Krishnamoorthi_Whitepaper_arXiv18}

\bibitem[{{Krithivasan} et~al.(2019){Krithivasan}, {Sen}, {Venkataramani}, and
  {Raghunathan}}]{Ref_Krithivasan_SpikeBundle_ISLPED19}
{Krithivasan}, S., {Sen}, S., {Venkataramani}, S., and {Raghunathan}, A.
  (2019).
\newblock Dynamic spike bundling for energy-efficient spiking neural networks.
\newblock In \emph{2019 IEEE/ACM International Symposium on Low Power
  Electronics and Design}. 1--6.
\newblock \doi{10.1109/ISLPED.2019.8824897}
\bibAnnoteFile{Ref_Krithivasan_SpikeBundle_ISLPED19}

\bibitem[{{Lecun} et~al.(1998){Lecun}, {Bottou}, {Bengio}, and
  {Haffner}}]{Ref_Lecun_MNIST_IEEE98}
{Lecun}, Y., {Bottou}, L., {Bengio}, Y., and {Haffner}, P. (1998).
\newblock Gradient-based learning applied to document recognition.
\newblock \emph{Proc. IEEE} 86, 2278--2324.
\newblock \doi{10.1109/5.726791}
\bibAnnoteFile{Ref_Lecun_MNIST_IEEE98}

\bibitem[{Liu et~al.(2019)Liu, Tang, Li, Cai, Zhang, and
  Zhou}]{Ref_Liu_Edge_JPROC19}
Liu, F., Tang, G., Li, Y., Cai, Z., Zhang, X., and Zhou, T. (2019).
\newblock A survey on edge computing systems and tools.
\newblock \emph{Proceedings of the IEEE} 107, 1537--1562.
\newblock \doi{10.1109/JPROC.2019.2920341}
\bibAnnoteFile{Ref_Liu_Edge_JPROC19}

\bibitem[{Micikevicius et~al.(2018)Micikevicius, Narang, Alben, Diamos, Elsen,
  Garc{\'{\i}}a et~al.}]{Ref_Micikevicius_MixedPrecision_ICLR18}
Micikevicius, P., Narang, S., Alben, J., Diamos, G.~F., Elsen, E.,
  Garc{\'{\i}}a, D., et~al. (2018).
\newblock Mixed precision training.
\newblock In \emph{6th International Conference on Learning Representations}
\bibAnnoteFile{Ref_Micikevicius_MixedPrecision_ICLR18}

\bibitem[{Mittal(2016)}]{Ref_Mittal_ApproxComSurvey_CSUR16}
Mittal, S. (2016).
\newblock A survey of techniques for approximate computing.
\newblock \emph{ACM Computing Survey} 48.
\newblock \doi{10.1145/2893356}
\bibAnnoteFile{Ref_Mittal_ApproxComSurvey_CSUR16}

\bibitem[{Mozafari et~al.(2019)Mozafari, Ganjtabesh, Nowzari-Dalini, and
  Masquelier}]{Ref_Mozafari_SpykeTorch_FNINS19}
Mozafari, M., Ganjtabesh, M., Nowzari-Dalini, A., and Masquelier, T. (2019).
\newblock Spyketorch: Efficient simulation of convolutional spiking neural
  networks with at most one spike per neuron.
\newblock \emph{Frontiers in Neuroscience} 13, 625.
\newblock \doi{10.3389/fnins.2019.00625}
\bibAnnoteFile{Ref_Mozafari_SpykeTorch_FNINS19}

\bibitem[{Nabavi~Larimi et~al.(2021)Nabavi~Larimi, Salami, Unsal, Kestelman,
  Sarbazi-Azad, and Mutlu}]{Ref_Larimi_HBMvolt_DATE21}
Nabavi~Larimi, S.~S., Salami, B., Unsal, O.~S., Kestelman, A.~C., Sarbazi-Azad,
  H., and Mutlu, O. (2021).
\newblock Understanding power consumption and reliability of high-bandwidth
  memory with voltage underscaling.
\newblock In \emph{2021 Design, Automation \& Test in Europe Conference \&
  Exhibition}. 517--522.
\newblock \doi{10.23919/DATE51398.2021.9474024}
\bibAnnoteFile{Ref_Larimi_HBMvolt_DATE21}

\bibitem[{Olgun et~al.(2021)Olgun, Luna, Kanellopoulos, Salami, Hassan, Ergin
  et~al.}]{Ref_Olgun_PiDRAM_arXiv21}
Olgun, A., Luna, J.~G., Kanellopoulos, K., Salami, B., Hassan, H., Ergin, O.,
  et~al. (2021).
\newblock \uppercase{P}i\uppercase{DRAM}: A holistic end-to-end
  \uppercase{FPGA}-based framework for processing-in-\uppercase{DRAM}.
\newblock \emph{arXiv preprint arXiv:2111.00082}
\bibAnnoteFile{Ref_Olgun_PiDRAM_arXiv21}

\bibitem[{Park et~al.(2019)Park, Kim, Choe, and Yoon}]{Ref_Park_BurstSNN_DAC19}
Park, S., Kim, S., Choe, H., and Yoon, S. (2019).
\newblock Fast and efficient information transmission with burst spikes in deep
  spiking neural networks.
\newblock In \emph{2019 56th Annual Design Automation Conference}.
\newblock \doi{10.1145/3316781.3317822}
\bibAnnoteFile{Ref_Park_BurstSNN_DAC19}

\bibitem[{{Park} et~al.(2020){Park}, {Kim}, {Na}, and
  {Yoon}}]{Ref_Park_T2FSNN_DAC20}
{Park}, S., {Kim}, S., {Na}, B., and {Yoon}, S. (2020).
\newblock \uppercase{T2FSNN}: Deep spiking neural networks with
  time-to-first-spike coding.
\newblock In \emph{57th ACM/IEEE Design Automation Conference}. 1--6.
\newblock \doi{10.1109/DAC18072.2020.9218689}
\bibAnnoteFile{Ref_Park_T2FSNN_DAC20}

\bibitem[{Pfeiffer and Pfeil(2018)}]{Ref_Pfeiffer_DLSNN_FNINS18}
Pfeiffer, M. and Pfeil, T. (2018).
\newblock Deep learning with spiking neurons: Opportunities and challenges.
\newblock \emph{Frontiers in Neuroscience} 12.
\newblock \doi{10.3389/fnins.2018.00774}
\bibAnnoteFile{Ref_Pfeiffer_DLSNN_FNINS18}

\bibitem[{{Putra} et~al.(2020){Putra}, {Hanif}, and
  {Shafique}}]{Ref_Putra_DRMap_DAC20}
{Putra}, R. V.~W., {Hanif}, M.~A., and {Shafique}, M. (2020).
\newblock \uppercase{DRM}ap: A generic \uppercase{DRAM} data mapping policy for
  energy-efficient processing of convolutional neural networks.
\newblock In \emph{2020 57th ACM/IEEE Design Automation Conference}. 1--6.
\newblock \doi{10.1109/DAC18072.2020.9218672}
\bibAnnoteFile{Ref_Putra_DRMap_DAC20}

\bibitem[{Putra et~al.(2021{\natexlab{a}})Putra, Hanif, and
  Shafique}]{Ref_Putra_ReSpawn_ICCAD21}
Putra, R. V.~W., Hanif, M.~A., and Shafique, M. (2021{\natexlab{a}}).
\newblock Re\uppercase{S}pawn: Energy-efficient fault-tolerance for spiking
  neural networks considering unreliable memories.
\newblock In \emph{2021 IEEE/ACM International Conference On Computer Aided
  Design}. 1--9.
\newblock \doi{10.1109/ICCAD51958.2021.9643524}
\bibAnnoteFile{Ref_Putra_ReSpawn_ICCAD21}

\bibitem[{Putra et~al.(2021{\natexlab{b}})Putra, Hanif, and
  Shafique}]{Ref_Putra_ROMANet_TVLSI21}
Putra, R. V.~W., Hanif, M.~A., and Shafique, M. (2021{\natexlab{b}}).
\newblock \uppercase{ROMAN}et: Fine-grained reuse-driven off-chip memory access
  management and data organization for deep neural network accelerators.
\newblock \emph{IEEE Transactions on Very Large Scale Integration (VLSI)
  Systems} 29, 702--715.
\newblock \doi{10.1109/TVLSI.2021.3060509}
\bibAnnoteFile{Ref_Putra_ROMANet_TVLSI21}

\bibitem[{Putra et~al.(2022)Putra, Hanif, and
  Shafique}]{Ref_Putra_SoftSNN_arXiv22}
Putra, R. V.~W., Hanif, M.~A., and Shafique, M. (2022).
\newblock \uppercase{S}oft\uppercase{SNN}: Low-cost fault tolerance for spiking
  neural network accelerators under soft errors.
\newblock \emph{arXiv preprint arXiv:2203.05523}
\bibAnnoteFile{Ref_Putra_SoftSNN_arXiv22}

\bibitem[{{Putra} and {Shafique}(2020)}]{Ref_Putra_FSpiNN_TCAD20}
{Putra}, R. V.~W. and {Shafique}, M. (2020).
\newblock \uppercase{FS}pi\uppercase{NN}: An optimization framework for
  memory-efficient and energy-efficient spiking neural networks.
\newblock \emph{IEEE Transactions on Computer-Aided Design of Integrated
  Circuits and Systems} 39, 3601--3613.
\newblock \doi{10.1109/TCAD.2020.3013049}
\bibAnnoteFile{Ref_Putra_FSpiNN_TCAD20}

\bibitem[{Putra and Shafique(2021{\natexlab{a}})}]{Ref_Putra_QSpiNN_IJCNN21}
Putra, R. V.~W. and Shafique, M. (2021{\natexlab{a}}).
\newblock \uppercase{Q-S}pi\uppercase{NN}: A framework for quantizing spiking
  neural networks.
\newblock In \emph{2021 International Joint Conference on Neural Networks}.
  1--8.
\newblock \doi{10.1109/IJCNN52387.2021.9534087}
\bibAnnoteFile{Ref_Putra_QSpiNN_IJCNN21}

\bibitem[{Putra and Shafique(2021{\natexlab{b}})}]{Ref_Putra_SpikeDyn_DAC21}
Putra, R. V.~W. and Shafique, M. (2021{\natexlab{b}}).
\newblock \uppercase{S}pike\uppercase{D}yn: A framework for energy-efficient
  spiking neural networks with continual and unsupervised learning capabilities
  in dynamic environments.
\newblock In \emph{2021 58th ACM/IEEE Design Automation Conference}.
  1057--1062.
\newblock \doi{10.1109/DAC18074.2021.9586281}
\bibAnnoteFile{Ref_Putra_SpikeDyn_DAC21}

\bibitem[{Putra and
  Shafique(2022{\natexlab{a}})}]{Ref_Putra_lpSpikeCon_arXiv22}
Putra, R. V.~W. and Shafique, M. (2022{\natexlab{a}}).
\newblock lp\uppercase{S}pike\uppercase{C}on: Enabling low-precision spiking
  neural network processing for efficient unsupervised continual learning on
  autonomous agents.
\newblock \emph{arXiv preprint arXiv:2205.12295}
\bibAnnoteFile{Ref_Putra_lpSpikeCon_arXiv22}

\bibitem[{Putra and Shafique(2022{\natexlab{b}})}]{Ref_Putra_tinySNN_arXiv22}
Putra, R. V.~W. and Shafique, M. (2022{\natexlab{b}}).
\newblock tiny\uppercase{SNN}: Towards memory-and energy-efficient spiking
  neural networks.
\newblock \emph{arXiv preprint arXiv:2206.08656}
\bibAnnoteFile{Ref_Putra_tinySNN_arXiv22}

\bibitem[{Rahimi~Azghadi et~al.(2014)Rahimi~Azghadi, Iannella, Al-Sarawi,
  Indiveri, and Abbott}]{Ref_Azghadi_Plasticity_JPROC14}
Rahimi~Azghadi, M., Iannella, N., Al-Sarawi, S.~F., Indiveri, G., and Abbott,
  D. (2014).
\newblock Spike-based synaptic plasticity in silicon: Design, implementation,
  application, and challenges.
\newblock \emph{Proceedings of the IEEE} 102, 717--737.
\newblock \doi{10.1109/JPROC.2014.2314454}
\bibAnnoteFile{Ref_Azghadi_Plasticity_JPROC14}

\bibitem[{{Rathi} et~al.(2019){Rathi}, {Panda}, and
  {Roy}}]{Ref_Rathi_PruneQuantizeSNN_TCAD18}
{Rathi}, N., {Panda}, P., and {Roy}, K. (2019).
\newblock \uppercase{STDP}-based pruning of connections and weight quantization
  in spiking neural networks for energy-efficient recognition.
\newblock \emph{IEEE Transactions on Computer-Aided Design of Integrated
  Circuits and Systems} 38, 668--677.
\newblock \doi{10.1109/TCAD.2018.2819366}
\bibAnnoteFile{Ref_Rathi_PruneQuantizeSNN_TCAD18}

\bibitem[{{Roy} et~al.(2017){Roy}, {Venkataramani}, {Gala}, {Sen},
  {Veezhinathan}, and {Raghunathan}}]{Ref_Roy_PEASE_ISLPED17}
{Roy}, A., {Venkataramani}, S., {Gala}, N., {Sen}, S., {Veezhinathan}, K., and
  {Raghunathan}, A. (2017).
\newblock A programmable event-driven architecture for evaluating spiking
  neural networks.
\newblock In \emph{2017 IEEE/ACM International Symposium on Low Power
  Electronics and Design}. 1--6.
\newblock \doi{10.1109/ISLPED.2017.8009176}
\bibAnnoteFile{Ref_Roy_PEASE_ISLPED17}

\bibitem[{Satyanarayanan(2017)}]{Ref_Satyanarayanan_Edge_MC17}
Satyanarayanan, M. (2017).
\newblock The emergence of edge computing.
\newblock \emph{Computer} 50, 30--39.
\newblock \doi{10.1109/MC.2017.9}
\bibAnnoteFile{Ref_Satyanarayanan_Edge_MC17}

\bibitem[{{Saunders} et~al.(2018){Saunders}, {Siegelmann}, {Kozma}, and
  {Ruszink´o}}]{Ref_Saunders_STDPpatch_IJCNN18}
{Saunders}, D.~J., {Siegelmann}, H.~T., {Kozma}, R., and {Ruszink´o}, M.
  (2018).
\newblock \uppercase{STDP} learning of image patches with convolutional spiking
  neural networks.
\newblock In \emph{2018 International Joint Conference on Neural Networks}.
  1--7.
\newblock \doi{10.1109/IJCNN.2018.8489684}
\bibAnnoteFile{Ref_Saunders_STDPpatch_IJCNN18}

\bibitem[{{Sen} et~al.(2017){Sen}, {Venkataramani}, and
  {Raghunathan}}]{Ref_Sen_ApproxSNN_DATE17}
{Sen}, S., {Venkataramani}, S., and {Raghunathan}, A. (2017).
\newblock Approximate computing for spiking neural networks.
\newblock In \emph{Design, Automation Test in Europe Conf. Exhibition}.
  193--198.
\newblock \doi{10.23919/DATE.2017.7926981}
\bibAnnoteFile{Ref_Sen_ApproxSNN_DATE17}

\bibitem[{Shafique et~al.(2018)Shafique, Khalid, and
  Rehman}]{Ref_Shafique_SmartCPS_DSD18}
Shafique, M., Khalid, F., and Rehman, S. (2018).
\newblock Intelligent security measures for smart cyber physical systems.
\newblock In \emph{2018 21st Euromicro Conference on Digital System Design}.
  280--287.
\newblock \doi{10.1109/DSD.2018.00058}
\bibAnnoteFile{Ref_Shafique_SmartCPS_DSD18}

\bibitem[{Shafique et~al.(2021)Shafique, Marchisio, Putra, and
  Hanif}]{Ref_Shafique_EdgeAI_ICCAD21}
Shafique, M., Marchisio, A., Putra, R. V.~W., and Hanif, M.~A. (2021).
\newblock \uppercase{T}owards energy-efficient and secure edge \uppercase{AI}:
  A cross-layer framework \uppercase{ICCAD} special session paper.
\newblock In \emph{2021 IEEE/ACM International Conference On Computer Aided
  Design}. 1--9.
\newblock \doi{10.1109/ICCAD51958.2021.9643539}
\bibAnnoteFile{Ref_Shafique_EdgeAI_ICCAD21}

\bibitem[{Shi et~al.(2016)Shi, Cao, Zhang, Li, and Xu}]{Ref_Shi_Edge_JIOT16}
Shi, W., Cao, J., Zhang, Q., Li, Y., and Xu, L. (2016).
\newblock Edge computing: Vision and challenges.
\newblock \emph{IEEE Internet of Things Journal} 3, 637--646.
\newblock \doi{10.1109/JIOT.2016.2579198}
\bibAnnoteFile{Ref_Shi_Edge_JIOT16}

\bibitem[{Strubell et~al.(2019)Strubell, Ganesh, and
  McCallum}]{Ref_Strubell_Carbon_ACL19}
Strubell, E., Ganesh, A., and McCallum, A. (2019).
\newblock Energy and policy considerations for deep learning in {NLP}.
\newblock In \emph{Proc. 57th Annual Meeting of the Association for
  Computational Linguistics}. 3645--3650.
\newblock \doi{10.18653/v1/P19-1355}
\bibAnnoteFile{Ref_Strubell_Carbon_ACL19}

\bibitem[{Strubell et~al.(2020)Strubell, Ganesh, and
  McCallum}]{Ref_Strubell_Carbon_AAAI20}
Strubell, E., Ganesh, A., and McCallum, A. (2020).
\newblock Energy and policy considerations for modern deep learning research.
\newblock \emph{Proc. AAAI Conference on Artificial Intelligence} 34,
  13693--13696.
\newblock \doi{10.1609/aaai.v34i09.7123}
\bibAnnoteFile{Ref_Strubell_Carbon_AAAI20}

\bibitem[{{Sze} et~al.(2017){Sze}, {Chen}, {Yang}, and
  {Emer}}]{Ref_Sze_DNNsurvey_IEEE17}
{Sze}, V., {Chen}, Y., {Yang}, T., and {Emer}, J.~S. (2017).
\newblock Efficient processing of deep neural networks: A tutorial and survey.
\newblock \emph{Proc. IEEE} 105, 2295--2329.
\newblock \doi{10.1109/JPROC.2017.2761740}
\bibAnnoteFile{Ref_Sze_DNNsurvey_IEEE17}

\bibitem[{Tavanaei et~al.(2019)Tavanaei, Ghodrati, Kheradpisheh, Masquelier,
  and Maida}]{Ref_Tavanaei_DLSNN_Neunet18}
Tavanaei, A., Ghodrati, M., Kheradpisheh, S.~R., Masquelier, T., and Maida, A.
  (2019).
\newblock Deep learning in spiking neural networks.
\newblock \emph{Neural Networks} 111, 47--63.
\newblock \doi{10.1016/j.neunet.2018.12.002}
\bibAnnoteFile{Ref_Tavanaei_DLSNN_Neunet18}

\bibitem[{Thorpe and Gautrais(1998)}]{Ref_Thorpe_RankOrder_Springer98}
Thorpe, S. and Gautrais, J. (1998).
\newblock Rank order coding.
\newblock In \emph{Computational Neuroscience}. 113--118.
\newblock \doi{10.1007/978-1-4615-4831-7_19}
\bibAnnoteFile{Ref_Thorpe_RankOrder_Springer98}

\bibitem[{van Baalen et~al.(2022)van Baalen, Kahne, Mahurin, Kuzmin, Skliar,
  Nagel et~al.}]{Ref_vanBaalen_SimulatedQuant_ICCV22}
van Baalen, M., Kahne, B., Mahurin, E., Kuzmin, A., Skliar, A., Nagel, M.,
  et~al. (2022).
\newblock Simulated quantization, real power savings.
\newblock In \emph{Proceedings of the IEEE/CVF Conference on Computer Vision
  and Pattern Recognition}. 2757--2761
\bibAnnoteFile{Ref_vanBaalen_SimulatedQuant_ICCV22}

\bibitem[{Venkataramani et~al.(2015)Venkataramani, Chakradhar, Roy, and
  Raghunathan}]{Ref_Venkataramani_ApproxCom_DAC15}
Venkataramani, S., Chakradhar, S.~T., Roy, K., and Raghunathan, A. (2015).
\newblock Approximate computing and the quest for computing efficiency.
\newblock In \emph{2015 52nd ACM/EDAC/IEEE Design Automation Conference}. 1--6.
\newblock \doi{10.1145/2744769.2744904}
\bibAnnoteFile{Ref_Venkataramani_ApproxCom_DAC15}

\bibitem[{Xu et~al.(2016)Xu, Mytkowicz, and
  Kim}]{Ref_Xu_ApproxCompSurvey_MDAT16}
Xu, Q., Mytkowicz, T., and Kim, N.~S. (2016).
\newblock Approximate computing: A survey.
\newblock \emph{IEEE Design \& Test} 33, 8--22.
\newblock \doi{10.1109/MDAT.2015.2505723}
\bibAnnoteFile{Ref_Xu_ApproxCompSurvey_MDAT16}

\bibitem[{Xu et~al.(2020)Xu, Tanvir~Arafin, and Qu}]{Ref_Xu_MIDAS_AsianHOST20}
Xu, Q., Tanvir~Arafin, M., and Qu, G. (2020).
\newblock \uppercase{MIDAS}: Model inversion defenses using an approximate
  memory system.
\newblock In \emph{2020 Asian Hardware Oriented Security and Trust Symposium}.
  1--4.
\newblock \doi{10.1109/AsianHOST51057.2020.9358254}
\bibAnnoteFile{Ref_Xu_MIDAS_AsianHOST20}

\bibitem[{Yu et~al.(2018)Yu, Liang, He, Hatcher, Lu, Lin
  et~al.}]{Ref_Yu_Edge_Access18}
Yu, W., Liang, F., He, X., Hatcher, W.~G., Lu, C., Lin, J., et~al. (2018).
\newblock A survey on the edge computing for the internet of things.
\newblock \emph{IEEE Access} 6, 6900--6919.
\newblock \doi{10.1109/ACCESS.2017.2778504}
\bibAnnoteFile{Ref_Yu_Edge_Access18}

\end{thebibliography}

%%% Make sure to upload the bib file along with the tex file and PDF
%%% Please see the test.bib file for some examples of references

\end{document}